\documentclass{tmsce}

\usepackage{lipsum}
\usepackage{amsmath,amsfonts,amsthm}
\usepackage{algorithmic} % Note: Template uses algorithmic, you used algorithmicx.
\usepackage{algorithm}
\usepackage{array}
\usepackage[caption=false,font=normalsize,labelfont=sf,textfont=sf]{subfig}
\usepackage{textcomp}
\usepackage{stfloats}
\usepackage{url}
\usepackage{verbatim}
\usepackage{graphicx}
\usepackage{cite} % Template's citation package
\usepackage{mwe}
\usepackage[switch]{lineno} % 
\makeatletter
\def\makeLineNumberLeft{%
  \hss\linenumberfont\LineNumber\hskip\marginparsep
}
\def\makeLineNumberRight{%
  \hskip\marginparsep\linenumberfont\LineNumber\hss
}
\def\makeLineNumber{%
  \hb@xt@\z@{\makeLineNumberLeft}%
  \makebox[\textwidth][l]{}%
  \hb@xt@\z@{\makeLineNumberRight}% 
}
\makeatother
\usepackage{xcolor}
\hypersetup{
    colorlinks=true,
    linkcolor=blue,
    citecolor=blue,
    urlcolor=blue,
    bookmarksnumbered=true
}
\usepackage[title]{appendix}
\usepackage{mathrsfs}
\usepackage{gensymb}
\usepackage{booktabs}
\usepackage{threeparttable}
\usepackage{multirow}
\usepackage{enumitem}
\usepackage{chngcntr}
\usepackage{tabularx}
\usepackage{csquotes}
\usepackage{diagbox}
\usepackage{float}

\newcommand{\parencite}[1]{\cite{#1}}
\newcommand{\textcite}[1]{\cite{#1}}

\title{PLAS-Net: Pixel-Level Area Segmentation for UAV-Based Beach Litter Monitoring}

\authors{
Yongying Liu$^{a}$,
Jiaqi Wang$^{a,1}$,
Jian 
Song$^{a,1}$,
Xinlei Shao$^{a,1}$,
Yijia Chen$^{a}$,
Nan Xu$^{c}$,
Katsunori Mizuno$^{a}$,
Shigeru Tabeta$^{a}$,
and Fan Zhao$^{a,*}$
}

\affiliation{$^{a}$Graduate School of Frontier Sciences, The University of Tokyo, Japan\\
$^{c}$Department of Urban Informatics, Shenzhen University, China\\[4pt]
\textit{Corresponding author:}\\
\texttt{\nolinkurl{zhaofan25ut@163.com}} (F.~Zhao)\\
{\footnotesize $^{1}$These authors contributed equally to this work.}
}

\begin{document}
% \linenumbers % 
\maketitle

\begin{abstract}
Accurate quantification of the physical exposure area of beach litter, rather than simple item counts, is essential for credible ecological risk assessment of marine debris.
However, automated UAV-based monitoring predominantly relies on bounding-box detection, which systematically overestimates the planar area of irregular litter objects.
To address this geometric limitation, we develop PLAS-Net (Pixel-level Litter Area Segmentor), an instance segmentation framework that extracts pixel-accurate physical footprints of coastal debris.
Evaluated on UAV imagery from a monsoon-driven pocket beach in Koh Tao, Thailand, PLAS-Net achieves a $mAP_{50}$ of 58.7\% with higher precision than eleven baseline models, demonstrating improved mask fidelity under complex coastal conditions.
To illustrate how the accuracy of the masking affects the conclusions of environmental analysis, we conducted three downstream demonstrations: (i) power-law fitting of normalized plastic density (NPD) to characterize fragmentation dynamics; (ii) area-weighted ecological risk index (ERI) to map spatial pollution hotspots; and (iii) source composition analysis revealing the abundance-area paradox: fishing gear constitutes a small proportion of the total number of items, but has the largest physical area per unit item. Pixel-level area extraction can provide more valuable information for coastal monitoring compared to methods based solely on counting.

\end{abstract}

\printkeywords

% \printdates

\tmsceendfrontmatter % End of Introduction

% ==========================================================

% Main Text

% ==========================================================

\section{Introduction}

Anthropogenic marine debris, mainly composed of plastic waste, has become a persistent environmental threat to global coastal ecosystems. \parencite{macleodGlobalThreatPlastic2021, barnesAccumulationFragmentationPlastic2009}. Addressing the environmental and socioeconomic challenges of coastal areas requires scalable management solutions. Between 4.8 million and 12.7 million tons of plastic waste flow into the ocean annually, accumulating on coastlines and threatening marine biodiversity and tourism revenue. Understanding the distribution of marine debris is crucial for effectively mitigating its impact\parencite{ryanMonitoringAbundancePlastic2009}.

Traditional monitoring methods rely on artificial beach surveys, which are labor-intensive and limited by observer subjectivity and spatial coverage. Combining consumer-grade drones with deep learning algorithms has become a cost-effective and automated method for detecting marine debris\parencite{Maximenko2019_IMDOS}. Bounding-based architectures and amphibious drones have been deployed to detect floating debris\parencite{liu2026water}, riverbed and seabed debris\parencite{zhao2024riverbed, zhao2025seafloor, tao2025diffusion}, and shallow-water ecosystems\parencite{zhao2025novel, zhao2026cost}. These technologies are particularly useful in remote areas with limited management resources. Koh Tao in the Gulf of Thailand is far from the mainland and, influenced by seasonal monsoons, has become a topographical convergence point for cross-border marine pollution\parencite{Phattananuruch2024_JMSE_GoT_FMD, Yanagi2001_GoT_Monsoon}.

While UAV remote sensing has catalyzed a transition from manual surveys to automated monitoring, the methodological evolution remains constrained by a fundamental physical limitation.
Current object detection frameworks provide rapid abundance counts but remain constrained by the bounding box paradigm \parencite{martinUseUnmannedAerial2018, zhaoApplicationImprovedMachine2024}. Bounding boxes enclose substantial background areas, which overestimates the planar exposure surface of irregular beach litter. This geometric distortion invalidates polymer degradation models and obscures the ecological risk posed to benthic habitats \parencite{SmithTurrell2021, Pengsakun2026}. The transition from macro litter to secondary microplastics is governed by the surface area-to-volume ratio and photochemical cleavage \parencite{Andrady2011, cozarPlasticDebrisOpen2014}, making accurate area measurement a physical requirement rather than a methodological preference. Pixel-level mask extraction therefore provides a more appropriate geometric basis for fragmentation analysis.

Instance segmentation addresses this limitation by extracting pixel-level masks for individual beach litter objects \parencite{scarricaNovelBeachLitter2022, zouHighQualityInstanceSegmentationNetwork2022}. In optical remote sensing on sandy beaches, these masks represent the two-dimensional exposure footprint of the debris. It quantifies the physical interface that governs the rate of ultraviolet photodegradation and determines direct contact risk to coastal fauna \parencite{smithMonitoringPlasticBeach2021}. Acquiring this geometric data is a prerequisite for shifting from qualitative observation to quantitative physical degradation modeling. To address the geometric limitations of bounding box methods, this study develops the Pixel-level Litter Area Segmentor (PLAS-Net), an instance segmentation framework for coastal environments.

Recent UAV-based coastal monitoring studies have applied deep learning to reduce manual labor. Early approaches relied on bounding box detection algorithms to accelerate item counting. For instance, Fallati et al.
\cite{FALLATI2019133581} deployed deep learning detection architectures on UAV imagery across the beaches of the Republic of Maldives to automate macro-litter quantification.
However, these detection algorithms fail to capture the physical footprints ofdebris.
To address spatial metrics, subsequent studies transitioned to semantic segmentation. Notably, Song et al.
\cite{su14148311} utilized deep learning-based segmentation models to estimate the physical covered area of marine debris across an uninhabited coastal island in Korea.
Such approaches process pixels homogeneously, lacking the capacity to distinguish overlapping or packed individual instances.
Recently, conventional instance segmentation models have been introduced to overcome this limitation. For example, Sozio et al.
\cite{rs16193617} tested a Mask R-CNN-based approach on UAV imagery from the Torre Guaceto Marine Protected Area in Italy.
However, their findings indicated that the direct application of these conventional algorithms to natural, unstructured beaches remains problematic, highlighting the need for further architectural improvements.

Despite its ecological necessity, deploying instance segmentation in natural and unstructured beach environments introduces severe computer vision challenges \parencite{chengMethodsDatasetsSemantic2024}.
First, the complex background textures of sand, shells, and coastal vegetation induce feature confusion and severe background noise.
Second, the inevitable partial burial and occlusion of debris by sand frequently cause the predicted masks to fragment into disconnected patches. Finally, marine litter exhibits extreme scale variations where thin and elongated structures like fishing lines and deformed nets easily lose their topological continuity during deep network feature extraction \parencite{sapkotaYOLO26KeyArchitectural2026}.
These field-specific interferences severely degrade the performance of conventional segmentation models.
To simultaneously address these ecological quantification needs and complex visual deployment challenges, this study proposes PLAS-Net (Pixel-level Litter Area Segmentor), an advanced instance segmentation framework based on the YOLOv26n-seg architecture \parencite{sapkotaYOLO26KeyArchitectural2026}.
We introduce three customized modules explicitly designed to overcome the identified coastal vision bottlenecks.
Initially, a C3k2+Dynamic Feature Fusion (C3kDFF) module is proposed to effectively suppress background noise from complex beach textures.
Subsequently, a C2PSA+Context Anchor Attention (CPCAA) module is introduced to capture long-range dependencies and repair mask fragmentation caused by partial burial.
Lastly, a Dynamic Multi-scale Sequence Fusion (DMSSF) module is developed to maintain the cross-scale structural integrity of highly deformed and elongated fishing gear.

The contribution of this study is an instance segmentation framework, 
PLAS-Net, that produces more accurate pixel-level litter masks than existing 
bounding-box and segmentation baselines under complex coastal conditions.
To 
demonstrate the practical value of improved mask quality, we apply the resulting 
area measurements to three downstream environmental analyses: fragmentation 
dynamics modelled via normalized plastic density (NPD) power-law fitting 
\parencite{cozarPlasticDebrisOpen2014};
spatial pollution intensity mapped using 
an area-weighted Ecological Risk Index (ERI) \parencite{hakansonEcologicalRiskIndex1980};
and source composition analysis that decouples abundance from physical footprint 
to infer source-sink dynamics.
Together, these downstream applications illustrate 
how higher-fidelity geometric data can support more credible coastal monitoring and 
evidence-based conservation planning in monsoon-driven environments.The primary contribution of this study is an instance segmentation framework, PLAS-Net, that produces more accurate pixel-level litter masks than existing bounding-box and segmentation baselines under complex coastal conditions.
To demonstrate the practical value of improved mask quality, we apply the resulting area measurements to three downstream environmental analyses: fragmentation dynamics modelled via NPD power-law fitting \parencite{cozarPlasticDebrisOpen2014};
spatial pollution intensity mapped using an area-weighted ERI \parencite{hakansonEcologicalRiskIndex1980}; and source composition analysis that decouples abundance from physical footprint to infer source-sink dynamics.
Together, these downstream applications illustrate how higher-fidelity geometric data can support more credible coastal monitoring and evidence-based conservation planning in monsoon-driven environments.

\section{Methodology}

\begin{figure*}[htbp]
    \centering
    \includegraphics[width=\textwidth]{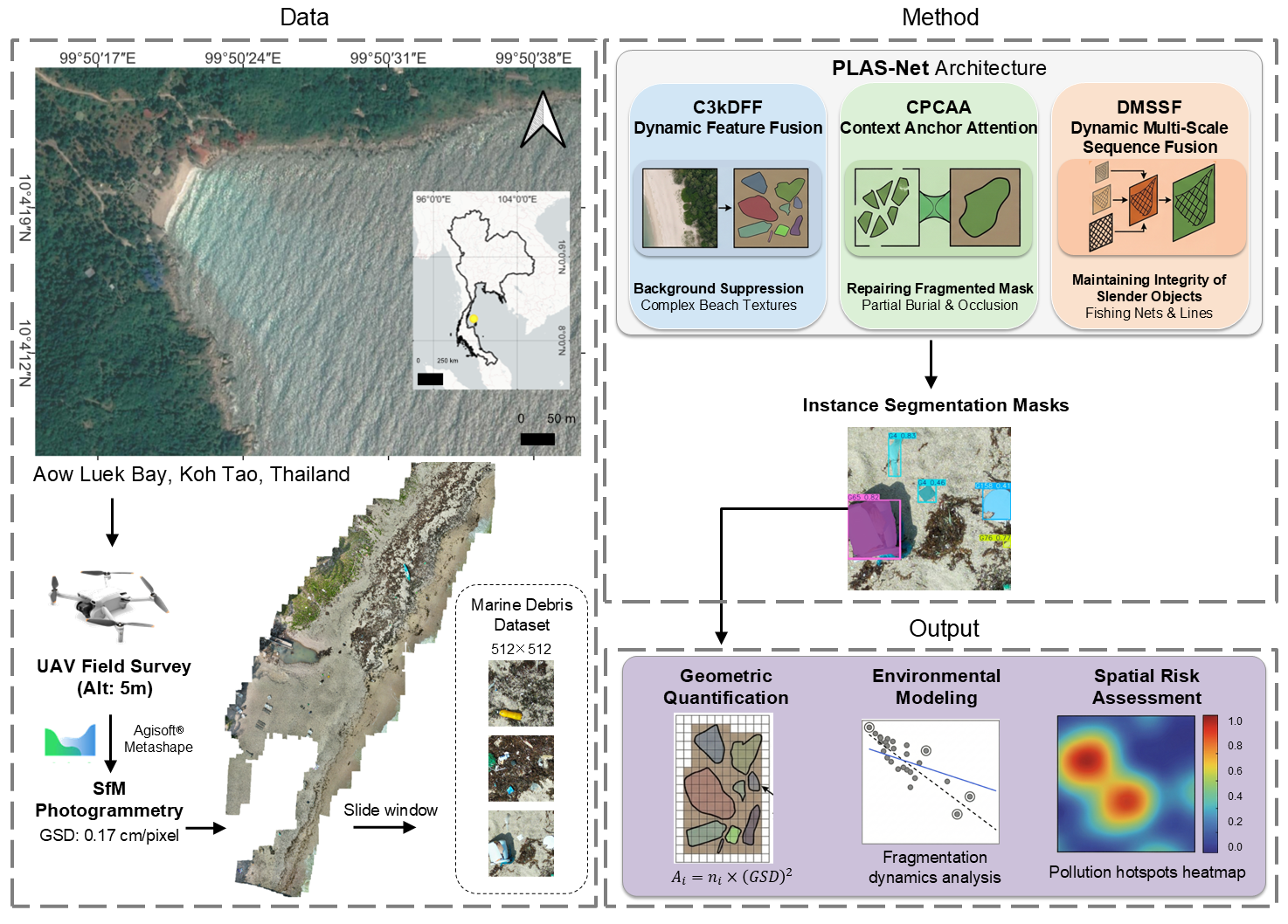}
    \caption{The workflow is organized into three tiers. \textbf{Phase 1: Data Acquisition} focuses on data generation, utilizing UAV-SfM reconstruction at Aow Luek Bay to produce high-resolution orthomosaics (GSD: $0.17$ cm/pixel) for $512 \times 512$-pixel dataset extraction. 
\textbf{Phase 2: Methodology} introduces the PLAS-Net architecture, specifically engineered with C3kDFF, CPCAA, and DMSSF modules to suppress background noise and rectify mask fragmentation for high-fidelity instance segmentation. 
\textbf{Phase 3: Output \& Analysis} bridges the gap between pixel-level masks and environmental assessment, extracting physical geometric footprints ($A_i$) to drive downstream models, including fragmentation dynamics (NPD) and ecological risk assessments.}
    \label{fig:overall_framework}
\end{figure*}

\subsection{Study Area and Litter Source Dynamics}

Aow Luek Bay, located on the southeast coast of Koh Tao in the western Gulf of Thailand (Fig. \ref{fig:overall_framework}), features a tropical semi-enclosed pocket beach laterally confined by granite headlands. This geomorphology enhances debris retention, as pocket beaches typically exhibit higher marine litter accumulation rates than open coastlines \parencite{CritchellLambrechts2016_ECSS_PlasticAccumulation, preveniosBeachLitterDynamics2018}. Although island-wide quantitative data are sparse, local observations identify Aow 
Luek Bay as a primary litter accumulation site on Koh Tao. Hydrodynamically, this accumulation is season-dependent.
During the northeast monsoon (November to April, encompassing the survey period), prevailing onshore winds and waves drive surface transport directly into the bay \parencite{Guo2021_GoT_NEM_Circulation, Phattananuruch2024_JMSE_GoT_FMD}.
Consequently, the bay traps both distant marine-derived waste transported via Gulf circulation \parencite{vansebillePhysicalOceanographyTransport2020} and local anthropogenic litter from tourism \parencite{lopez-arquilloInterdependenceCoastalTourist2024}.
Thus, Aow Luek Bay serves as an ideal site to investigate beach-scale litter distribution under mixed-source conditions.
\subsection{Data Acquisition and Preprocessing}

High-resolution RGB video imagery of the intertidal and supratidal zones at Ao Luek Bay was acquired in February 2024 using a DJI Mini 3 multi-rotor UAV.
Following standard operational procedures \parencite{martinUseUnmannedAerial2018, steinmetz-weissMappingDroneApplications2025}, flights were conducted at a 5-m altitude along the beach axis.
To ensure robust image matching for subsequent 3D reconstruction, forward and lateral overlaps were maintained at $\ge$ 75\% and $\ge$ 65\%, respectively, aligning with regional offshore monitoring protocols \parencite{Phattananuruch2024_JMSE_GoT_FMD}.
Surveys occurred during low tide (Figure~\ref{fig:overall_framework}) to maximize beach exposure and minimize interference from water specular reflection and wave-induced obstruction \parencite{preveniosBeachLitterDynamics2018}.
To ensure accurate geometric scaling for subsequent physical area estimation, the extracted UAV frames were processed using Agisoft Metashape software to generate a geometrically corrected orthomosaic, yielding a precise ground sampling distance (GSD) of approximately 0.17~cm/pixel.
This continuous orthomosaic was systematically cropped into $512 \times 512$ pixel tiles to fit the input dimensions of the network architecture \parencite{songComparativeStudyDeep2021} while preserving the fine-grained morphological details of irregular targets.
After filtering out tiles with severe motion blur, overexposure, or entirely lacking debris, the final dataset comprised 1,300 images (905 for training and 395 for testing, partitioned at the tile level). Because the orthomosaic was divided into non-overlapping $512 \times 512$ tiles prior to splitting, no individual litter object appears in both partitions.
Instance-level annotation was performed using LabelMe by two trained annotators working independently. Individual litter instances were manually delineated with polygon masks following a standardised annotation protocol developed prior to labelling. The protocol addressed three common difficult cases: (i) partially buried objects were bounded at the visible surface perimeter, excluding the estimated buried portion; (ii) thin and elongated items such as fishing lines and deformed nets were traced along their visible centreline with a minimum polygon width of two pixels; and (iii) ambiguous fragments smaller than $3 \times 3$ pixels at the $512 \times 512$ tile resolution were excluded to reduce label noise. A cross-check review was conducted on 15\% of tiles selected at random, with boundary disagreements exceeding 10\% IoU resolved by consensus. To maintain objectivity and global comparability, litter was classified into 14 categories based on the G-code system established by the EU Technical Group on Marine Litter \parencite{ClassifyPlasticWaste} (e.g., G4: plastic bags, G6: plastic bottles, G137: fishing line).
Furthermore, to decipher source-sink dynamics in subsequent analyses, these fine-grained categories were functionally aggregated into three macroscopic origins: Domestic (consumer goods and packaging), Fishing (maritime and aquacultural gear), and Fragments (unidentifiable degraded plastics).
\subsection{Overall Instance Segmentation Framework}
\label{sec:overall_framework}

\begin{figure}[H]
\centering
\includegraphics[width=\textwidth]{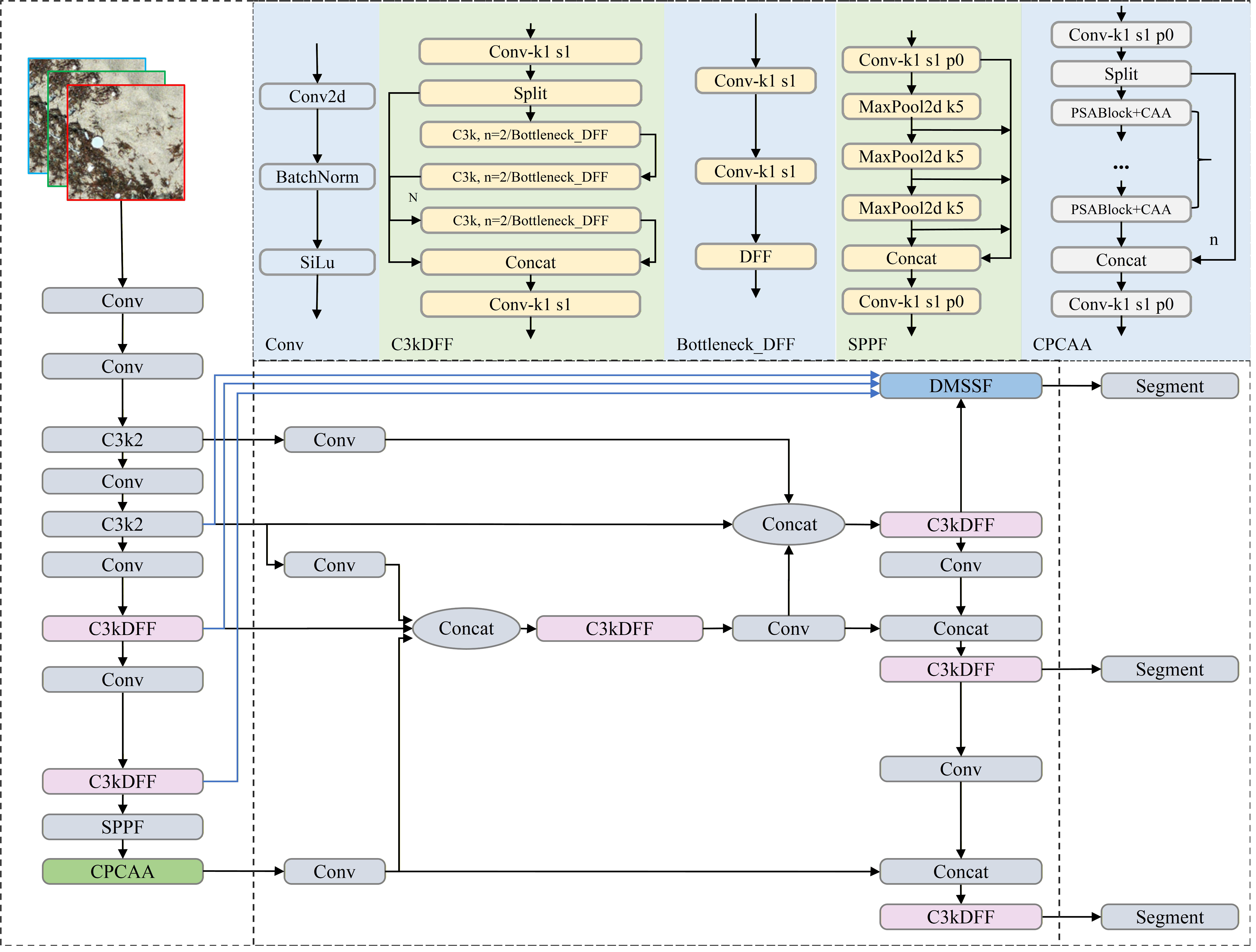}
\caption{Architecture of the PLAS-Net model integrating C3kDFF, CPCAA, and DMSSF modules.}
\label{fig:improved_yolov26n}
\end{figure}

To extract instance-level spatial and geometric information from UAV imagery, this study employed an instance segmentation approach.
Instance segmentation is defined as a computer vision technique that generates an independent pixel-level mask for each identified object, enabling the direct derivation of coverage area and morphological attributes.
According to the standard YOLOv26n-seg protocol, input images are processed through a convolutional backbone for feature extraction and a feature pyramid network for multi-scale fusion, followed by a segmentation head that predicts bounding boxes and pixel masks \parencite{sapkotaYOLO26KeyArchitectural2026}.
Despite the high inference speed of this baseline model, standard architectures frequently struggle with the unique imaging conditions of UAV-based beach surveys, such as severe background interference, target occlusion, and extreme scale variations.
To address these domain-specific challenges, we developed PLAS-Net, a framework tailored for beach litter detection.
Most importantly, this optimized framework provides pixel-level masks, which serve as the foundational data for the subsequent derivation of the NPD and the ERI.
As shown in Fig.~\ref{fig:improved_yolov26n}, the PLAS-Net architecture mainly consists of three integral components:

(i) a C3kDFF module, which adaptively calibrates channel and spatial features based on global context to suppress background noise from sand and vegetation;
(ii) a CPCAA module, which captures long-range spatial dependencies through directional receptive fields to repair mask fragmentation issues typically caused by partial burial or object overlap;
(iii) a DMSSF module, which integrates high-level semantics with low-level detailed features through three-dimensional sequential modeling to maintain the structural integrity of slender and irregular objects (e.g., fishing nets and ropes).

The following subsections detail these three modules and provide mathematical formulations for their respective feature enhancement mechanisms.
\subsubsection{C3kDFF for Background Suppression}
\label{sec:c3kdff}
\begin{figure}[H]
\centering
\includegraphics[width=0.8\textwidth]{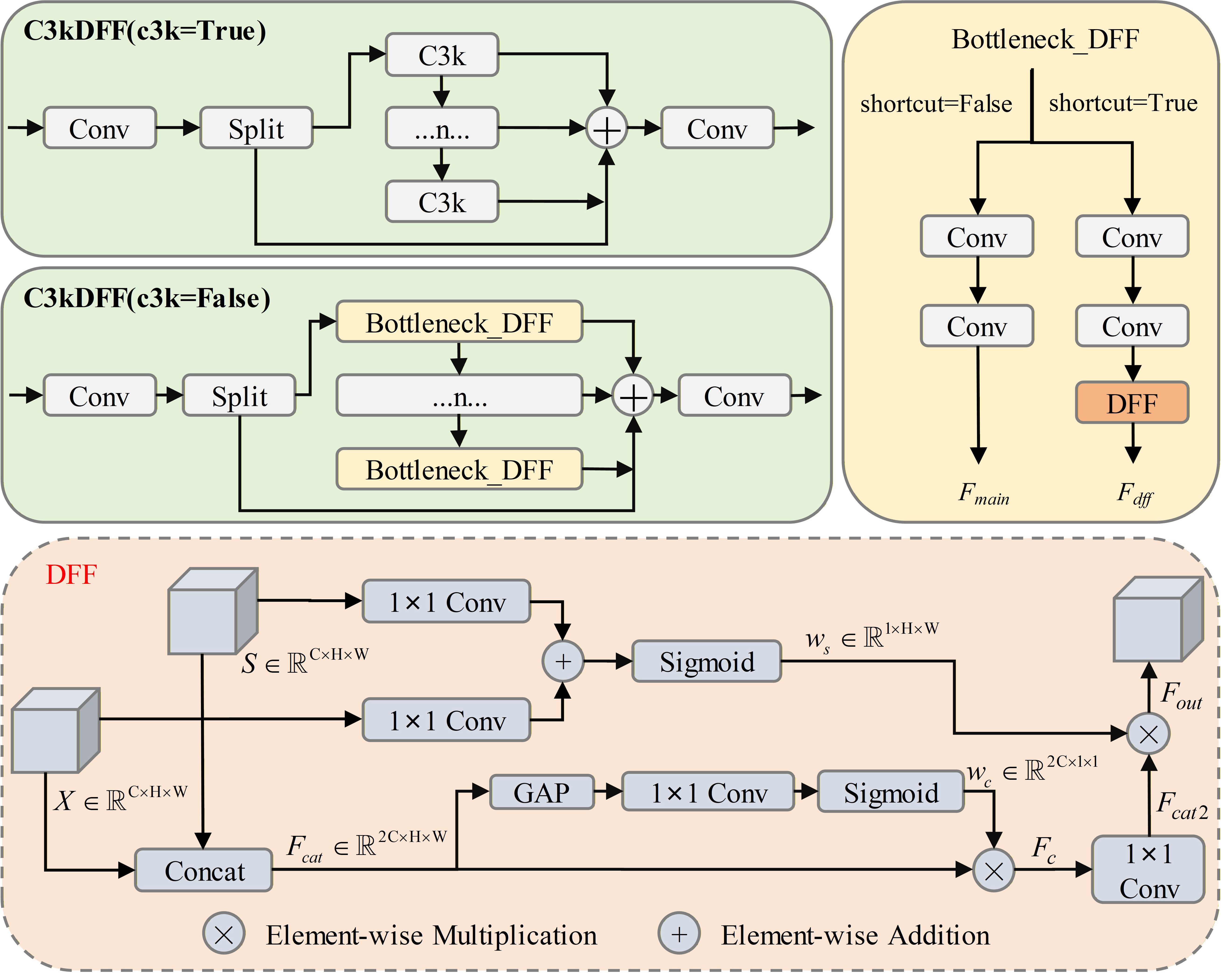}
\caption{Structural diagram of the proposed C3kDFF module.}
\label{fig:c3kdff}
\end{figure}

To address the loss of shallow texture details and poor localization of small, inconspicuous beach litter caused by static feature fusion \parencite{zhaoApplicationImprovedMachine2024, kangASFYOLONovelYOLO2024a}, we propose the C3kDFF module, drawing on the dynamic selection mechanism investigated by \textcite{yangDNetDynamicLarge2026}.
The C3kDFF module is a feature extraction unit that replaces the static residual connections of the baseline C3k2 structure with dynamic feature fusion to adaptively calibrate main and residual branch features using global context.
Figure~\ref{fig:c3kdff} illustrates the structure of this dynamic module.

Within the Bottleneck\_DFF unit, the DFF-equipped branch replaces the original two-convolution layer branch when input and output channel dimensions match.
Let $X$ and $S$ denote the main and residual branch features, respectively, with channel dimension $C$ and spatial dimensions $H \times W$.
For channel-level dynamic calibration, $X$ and $S$ are concatenated to obtain the fused feature $F$:
\begin{equation}
F = \mathrm{Concat}(X, S).
\end{equation}

Global Average Pooling (GAP) compresses $F$ into a channel descriptor.
A gating mechanism ($1 \times 1$ convolution and Sigmoid function $\sigma$) generates the channel-level dynamic weight $\alpha$:
\begin{equation}
\alpha = \sigma(\mathrm{Conv}_{1 \times 1}(\mathrm{GAP}(F))).
\end{equation}
This weight $\alpha$ calibrates $F$ to enhance critical channels and suppress background noise:
\begin{equation}
F' = \alpha \otimes F .
\end{equation}

A subsequent $1 \times 1$ convolution restores the channel dimension to $C$, yielding the channel-fused feature $F''$.
To further enhance litter region responses, a spatial dynamic selection mechanism generates a spatial weight $\beta$.
A $1 \times 1$ convolution compresses the channel dimensions of $X$ and $S$ to 1. These descriptors undergo element-wise addition ($\oplus$) and Sigmoid activation:
\begin{equation}
\beta = \sigma(\mathrm{Conv}_{1 \times 1}(X) \oplus \mathrm{Conv}_{1 \times 1}(S)).
\end{equation}
This weight $\beta$ encodes discriminative spatial regions focused on by both branches.
Finally, $F''$ is multiplied by $\beta$ for adaptive spatial enhancement:
\begin{equation}
Out = \beta \otimes F'' .
\end{equation}

The C3kDFF module replaces the C3k2 modules in the backbone (layers 6, 8) and neck (layers 14, 18, 21, 24).
This multi-level integration dynamically adjusts the fusion ratio of shallow details and deep semantics, preserving crucial geometric information under varying illuminations.
\subsubsection{CPCAA for Mask Fragmentation}
\label{sec:cpcaa}

\begin{figure}[H]
\centering
\includegraphics[width=0.7\textwidth]{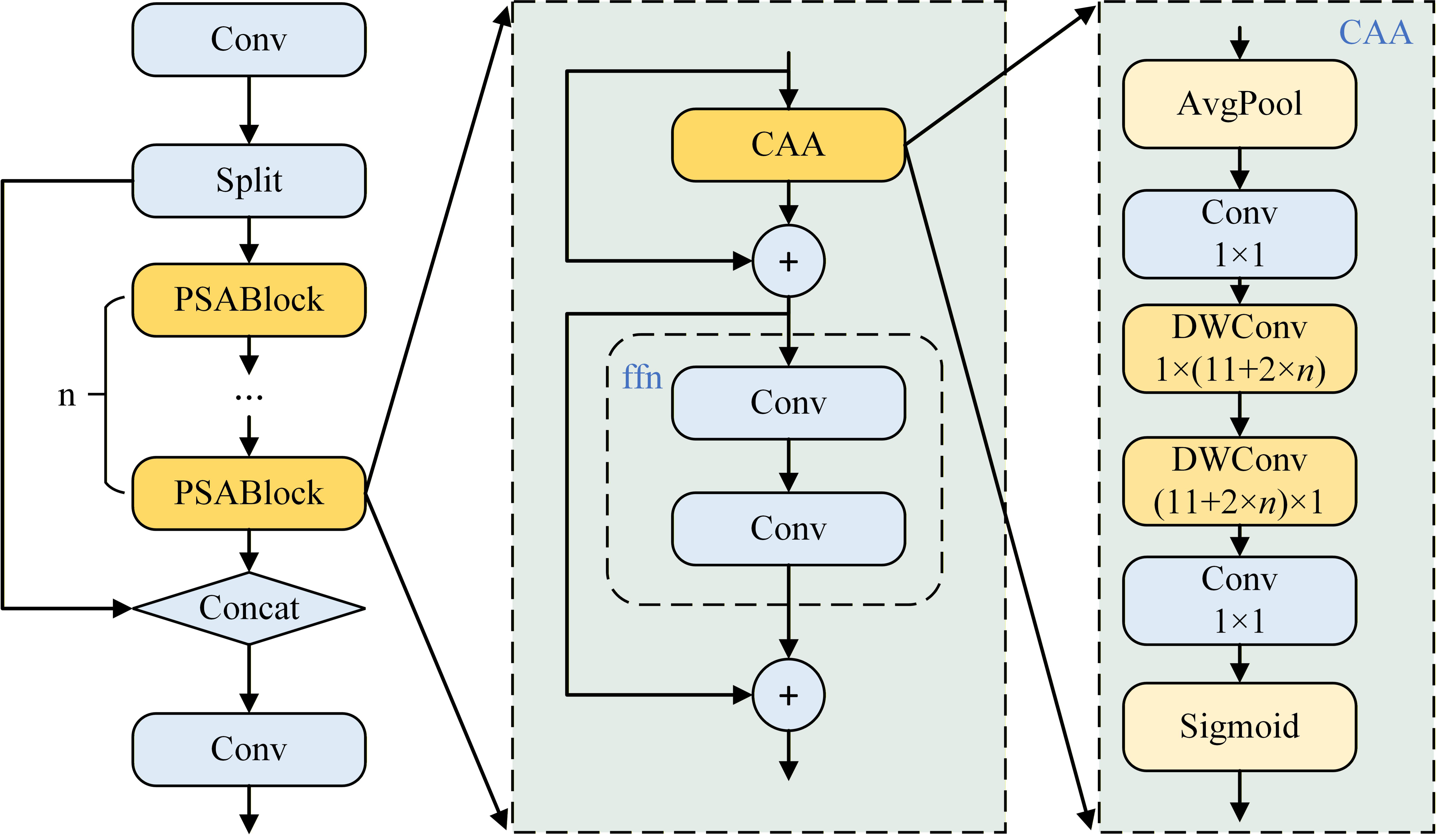}
\caption{Structure of the proposed CPCAA module.}
\label{fig:cpcaa}
\end{figure}

In coastal remote sensing, the instance segmentation of beach litter frequently encounters mask fragmentation caused by slender target morphologies and partial burial \parencite{scarricaNovelBeachLitter2022, hidakaPixellevelImageClassification2022, barryTop10Marine2023, zhaoApplicationImprovedMachine2024};
however, baseline modules such as C2PSA exhibit limited capability in capturing the strongly directional local context required to resolve these long-range dependencies.
To address this limitation, we introduce the CPCAA module, which incorporates the CAA mechanism examined by \textcite{caiPolyKernelInception2024a}.
The CAA is defined as an attention unit that models long-range spatial dependencies efficiently through global average pooling and one-dimensional strip convolutions.
Figure~\ref{fig:cpcaa} illustrates the structural details of this CPCAA module.

The CAA structure initially aggregates local neighborhood information through average pooling (AvgPool) and utilizes a $1 \times 1$ convolution for channel compression to extract global semantic context.
Subsequently, horizontal ($1 \times k$) and vertical ($k \times 1$) strip depth-wise convolutions form a directional receptive field.
This directional field captures long-range spatial dependencies of slender targets under lightweight computational constraints.
A subsequent $1 \times 1$ convolution and a Sigmoid function ($\sigma$) generate the spatial attention weight map.
This attention process is formally expressed as:
\begin{equation}
\mathrm{CAA}_{weight}(X) = \sigma(\mathrm{Conv}_{1 \times 1}(\mathrm{Conv}_{k \times 1}(\mathrm{Conv}_{1 \times k}(\mathrm{Conv}_{1 \times 1}(\mathrm{AvgPool}(X)))))).
\end{equation}

In the constructed CPCAA module, this CAA mechanism replaces the original multi-head attention block within the PSABlock structure.
The attention weights generated by the CAA are multiplied element-wise ($\otimes$) with the input feature $X$ to achieve feature enhancement for key contextual areas (e.g., litter accumulation zones):
\begin{equation}
Out = \mathrm{CAA}_{weight}(X) \otimes X .
\end{equation}

Compared to the traditional C2PSA module, this contextual enhancement enables the network to simultaneously learn fine-grained textures and global semantics.
This dual learning effectively repairs mask fragmentation issues typically observed when segmenting partially buried or heavily occluded beach debris.
\subsubsection{DMSSF for Slender Objects}
\label{sec:dmssf}

\begin{figure}[H]
\centering
\includegraphics[width=0.8\textwidth]{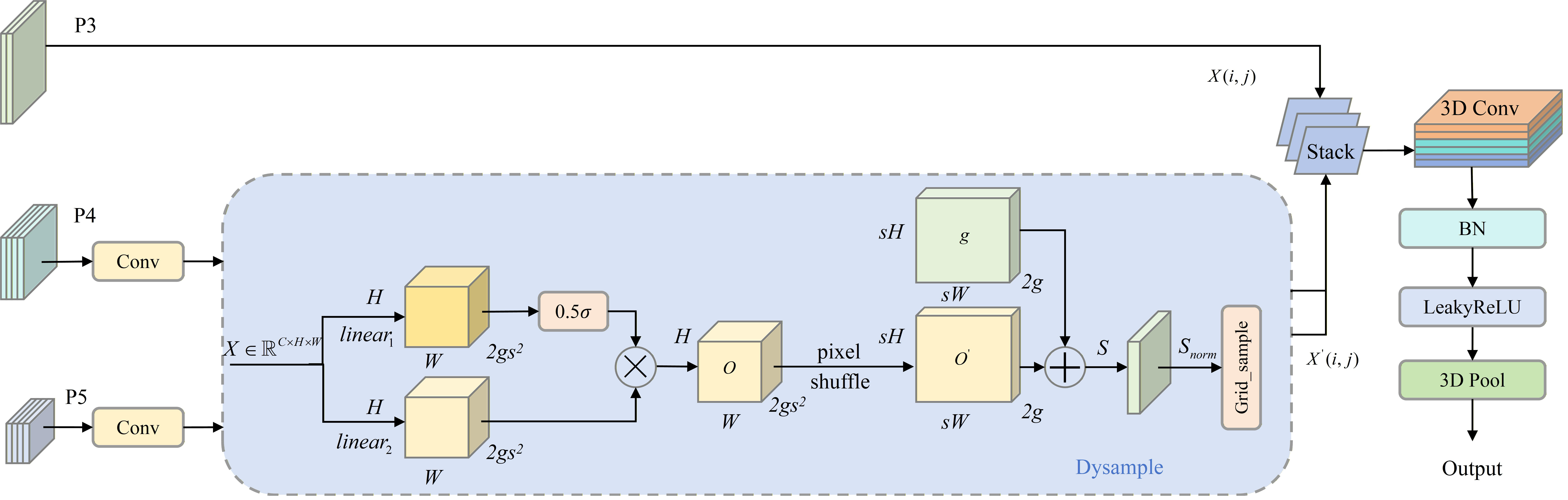}
\caption{Structure of the proposed DMSSF module.}
\label{fig:dmssf}
\end{figure}

Traditional feature pyramid structures widely utilized for multi-scale feature fusion \parencite{kangASFYOLONovelYOLO2024a} frequently fail to maintain the structural integrity of slender and irregular marine debris \parencite{barryTop10Marine2023}.
This mask fragmentation arises because conventional integration methods rely on simple addition or concatenation operations, struggling to fully exploit the deep semantic associations across spatial hierarchies.
To resolve this limitation, we introduce the DMSSF module. The DMSSF is defined as a feature fusion architecture that integrates high-level semantics with low-level detailed features through three-dimensional sequential modeling.
Figure~\ref{fig:dmssf} illustrates the structural details of this DMSSF module.

The DMSSF module initially unifies the channel dimensions of the multi-scale feature maps (P3, P4, and P5) using three independent $1 \times 1$ convolutions.
To leverage the high-resolution characteristics of the P3 layer for small-target details, the P4 and P5 feature maps are upsampled to the spatial dimensions of P3.
This process utilizes the DySample dynamic upsampling module \parencite{liuLearningUpsampleLearning2023}. DySample employs a point-sampling-based dynamic upsampling mechanism, which avoids computationally expensive dynamic convolutions.
Given the input feature map $X$, a linear projection generates the sampling offset $O$:
\begin{equation}
O = \mathrm{Linear}(X) \cdot \lambda ,
\end{equation}
where $\lambda$ denotes the offset scaling factor.
In dynamic mode, this scaling factor $\lambda$ is adaptively predicted from the features:
\begin{equation}
\lambda = \sigma(\mathrm{Linear}(X)) .
\end{equation}

The generated offset $O$ is reshaped into $O'$ via a Pixel Shuffle operation and added to the original sampling grid $G$ to obtain the final sampling point set $S$:
\begin{equation}
S = G + O' .
\end{equation}

The sampling point coordinates are subsequently normalized according to the spatial size:
\begin{equation}
S_{norm} = 2 \cdot \frac{S}{size} - 1 .
\end{equation}

Finally, a bilinear interpolation resampling operation calculates the upsampled feature map $Y$ based on the normalized coordinates:
\begin{equation}
Y = \mathrm{grid\_sample}(X, S_{norm}) .
\end{equation}

Unlike traditional interpolation, this dynamic sampling adaptively adjusts positions based on feature content, effectively preserving morphological details.
The processed features (P3, P'$4$, P'$5$) are then stacked along a newly added scale dimension to construct a three-dimensional feature tensor.
A 3D convolution reorganizes these scale sequence features:
\begin{equation}
Out = \mathrm{MaxPool}_{3D}(\mathrm{LeakyReLU}(\mathrm{BN}(\mathrm{Conv}_{3D}(\mathrm{Stack}(P_3, P'_4, P'_5))))) .
\end{equation}

Through 3D convolution, Batch Normalization (BN), and LeakyReLU activation, this 3D fusion extracts cross-scale local associations and contextual dependencies simultaneously.
By treating the hierarchical feature maps as a continuous scale sequence, this mechanism effectively preserves the structural continuity of slender objects across multiple receptive fields.
\subsection{Quantitative Assessment Metrics}

\subsubsection{Instance-Level Geometric and Morphological Quantification}
Accurate quantification of marine debris morphology is essential to assess its environmental burden, as discrete counts obscure dimensional variability \parencite{hidakaPixellevelImageClassification2022}.
To capture these spatial complexities, physical attributes were extracted from the segmentation masks.
The physical coverage area $A_i$ for the $i$-th litter instance was calculated by Eq.~\eqref{eq:area_calc}:
\begin{equation}
A_i = n_i \times (\mathrm{GSD})^2,
\label{eq:area_calc}
\end{equation}
where $n_i$ denotes the number of pixels within the $i$-th instance mask, and $\mathrm{GSD}$ represents the ground sampling distance.
\subsubsection{NPD and Fragmentation Dynamics}
Marine plastic pollution threatens ecosystems through continuous fragmentation, which increases bioavailability \parencite{macleodGlobalThreatPlastic2021}.
To investigate degradation driven by coastal morphodynamics, the surveyed beach was delineated into the intertidal and backshore zones.
Within each zone, the size-frequency distribution of detected objects was constructed using logarithmic binning to accommodate the extreme scale variations in debris area ($10^{-4}$ to $10^{1}$ $m^2$).
Following the established protocol for marine plastic fragmentation analysis \parencite{cozarPlasticDebrisOpen2014, smithMonitoringPlasticBeach2021}, the area range was partitioned into $M=14$ intervals with bin limits following a geometric progression.
This binning resolution provides an balance between size-frequency resolution and the statistical occupancy of individual bins, ensuring sufficient data points across five orders of magnitude for power-law fitting while avoiding excessive empty bins in the large-size tail \parencite{SmithTurrell2021, cozarPlasticDebrisOpen2014}.
Following the MSFD size classification \parencite{galgani2023guidance}, the boundary between the Macro and Meso ranges was set at $6.25 \times 10^{-4}$~m$^2$ (equivalent to $25 \times 25$~mm$^2$), corresponding to the conventional macro-litter threshold of 25~mm in the longest dimension. Power-law fitting was performed separately for the Macro and Meso segments, as single-range fitting across the instrument detection floor is theoretically unsupported: Cózar et al. observed that the power-law assumption held only above 5~mm in oceanic trawl samples, with systematic deviation below that threshold attributed to sampling truncation \parencite{cozarPlasticDebrisOpen2014}.
For each bin, the bin width was defined as $\Delta S_i = S_{i,high} - S_{i,low}$.
To ensure a linear representation in log-log space, the characteristic size of each bin ($S_i$) was defined as the geometric mean of its boundaries:
\begin{equation}
S_i = \sqrt{S_{i,low} \times S_{i,high}}.
\label{eq:bin_center}
\end{equation}

The NPD for each size class was then computed by dividing the abundance of litter instances ($n_i$) by the corresponding bin width:
\begin{equation}
\mathrm{NPD}(S_i) = \frac{n_i}{\Delta S_i},
\label{eq:npd_calc_detail}
\end{equation}
where this normalization ensures that the distribution density is independent of the binning interval scale, providing a consistent measure across different size scales.
The fragmentation pattern was subsequently evaluated by fitting the NPD to a power-law function:
\begin{equation}
\mathrm{NPD}(S) = c S^\alpha,
\label{eq:npd_power_law}
\end{equation}
where $c$ is a normalization constant and $\alpha$ is the scaling exponent derived via Ordinary Least Squares regression on log-transformed variables: $\log_{10}(\mathrm{NPD}) = \log_{10}(c) + \alpha \log_{10}(S)$.

\subsubsection{Ecological Risk and Spatial Mismatch Analysis}
Translating visual survey data into coastal management strategies requires identifying pollution hotspots where ecological exposure peaks \parencite{jambeckPlasticWasteInputs2015a}.
To systematically assess localized risks, the surveyed beach was discretized into consecutive alongshore spatial units. The beach transect was partitioned into ten equal-quantile latitudinal sectors (S1-S10, ordered from south to north) to ensure comparable litter instance counts per sector and to capture along-shore distributional gradients.
The CCI \parencite{alkalayCleancoastIndexNew2007} was calculated for the $j$-th unit to evaluate pollution abundance:
\begin{equation}
\mathrm{CCI}_j = \frac{N_j}{L_j},
\label{eq:cci}
\end{equation}
where $N_j$ denotes the total number of litter instances within the unit, and $L_j$ represents the unit length.
\begin{table}[htbp]
\centering
\caption{Hazard weights ($w_k$) assigned to each debris category following the expert-elicitation ranking of \textcite{wilcoxUsingExpertElicitation2016}. Wilcox et al.\ report ordinal ranks for entanglement and ingestion risk across marine taxa; ranks were mapped to a 1--10 integer scale (rank 1--4 $\to$ 9--10; 5--8 $\to$ 7--8; 9--12 $\to$ 5--6; 13--16 $\to$ 3--4; 17--20 $\to$ 1--2) for use in the ERI formula.}

\label{tab:hazard_weights}
\begin{tabular}{llcc}
\toprule
G-code & Description & Category & $w_k$ \\
\midrule
G4   & Plastic bags (carrier/bin-liner)        & Domestic items  & 8 \\
G6   & Plastic drink bottles ($\leq$0.5\,L)    & Domestic items  & 2 \\
G7   & Plastic drink bottles ($>$0.5\,L)       & Domestic items  & 2 \\
G18  & Plastic crates/boxes (fishing-related)  & Fishing gear    & 7 \\
G21  & Plastic caps and lids                   & Domestic items  & 5 \\
G65  & Plastic barrels/drums (fishing-related) & Fishing gear    & 7 \\
G76  & Plastic fragments 2.5--50\,cm           & Fragments       & 3 \\
G77  & Plastic/polystyrene fragments $>$50\,cm & Fragments       & 3 \\
G137 & Clothing and fabric pieces              & Domestic items  & 2 \\
G138 & Shoes and footwear                      & Domestic items  & 2 \\
G151 & Paper bags                              & Domestic items  & 1 \\
G158 & Other paper items                       & Domestic items  & 1 \\
G173 & Other wooden items                      & Domestic items  & 1 \\
\bottomrule
\end{tabular}
\end{table}

Because count-based indices inflate the severity of micro-fragmented debris while undervaluing intact objects, an area-based ERI was developed to integrate spatial footprints with the classical risk framework \parencite{hakansonEcologicalRiskIndex1980}:
\begin{equation}
\mathrm{ERI}_j = \sum_{k=1}^{K} w_k \times A_{jk},
\label{eq:eri}
\end{equation}
where $A_{jk}$ denotes the cumulative physical area of litter category $k$ within unit $j$, and $w_k$ denotes the category-specific ecological hazard weight. Hazard weights $w_k$ (Table~\ref{tab:hazard_weights}) were assigned by mapping the ordinal threat rankings from Wilcox et al. \parencite{wilcoxUsingExpertElicitation2016} Table~2 onto a 1 to 10 integer scale. Fishing-origin items (G18, G65) received elevated weights ($w_k = 7$) because Wilcox et al. ranked fishing gear categories (buoys, traps, nets) as the highest-threat items across all three taxa (seabirds, sea turtles, marine mammals). Plastic bags (G4, $w_k = 8$) were ranked second-highest, consistent with their rank of 5.7 in the Wilcox composite ranking. Unidentifiable plastic fragments (G76, G77, $w_k = 3$) were assigned lower weights reflecting their rank of 16.3 out of 20 in the same study.
Both CCI and ERI outputs were standardized using Min-Max normalization.
To quantify the spatial divergence between litter abundance and ecological risk, a weighted centroid shift analysis was conducted.
The geographic centroid based on count ($C_{\mathrm{count}}$) is defined as the unweighted mean of instance coordinates, whereas the risk-weighted centroid ($C_{\mathrm{ERI}}$) integrates the physical area and ecological hazard weight:
\begin{equation}
C_{\mathrm{ERI}} = \left( \frac{\sum_{i=1}^{N} A_i w_i x_i}{\sum_{i=1}^{N} A_i w_i}, \frac{\sum_{i=1}^{N} A_i w_i y_i}{\sum_{i=1}^{N} A_i w_i} \right),
\label{eq:centroid}
\end{equation}
where $(x_i, y_i)$ represents the projected coordinates of the $i$-th instance centroid.
The Euclidean distance $\delta = \| C_{\mathrm{ERI}} - C_{\mathrm{count}} \|_2$ measures spatial mismatch, identifying regions where high-risk items accumulate despite low numerical abundance. These metrics form an integrated analytical framework. Pixel-level geometric extractions are transformed into topological fragmentation models and weighted risk indices, providing the computational basis to decouple source-sink dynamics and assess environmental impacts in the subsequent discussion.

\section{Results}
\subsection{Training Protocol and Experimental Environment}

The computational environment and training configurations are described here to support reproducibility and future benchmarking. All experiments were run on a high-performance workstation, with specifications listed in Table~\ref{tab:environment}. The hardware included an Intel(R) Xeon(R) Platinum 8470Q CPU with 25 vCPUs and 90 GB of system memory. An NVIDIA GeForce RTX 5090 GPU with 32 GB of VRAM handled deep learning computations. The software 
environment was built on Ubuntu 22.04, with PyTorch 2.7.0 and CUDA 12.8 used for tensor operations.
\begin{table}[ht]
\centering
\caption{Summary of the experimental environment parameters.}
\label{tab:environment}
\begin{tabular}{ll}
\hline
Parameter & Value \\ \hline
Operating System & Ubuntu 22.04 \\
System Memory & 90 GB \\
Graphics Card (GPU) & NVIDIA GeForce RTX 5090 (32 GB) \\
CPU & Intel(R) Xeon(R) Platinum 8470Q \\
Framework & PyTorch 2.7.0 \\
CUDA Version & 12.8 \\
Python Version & 3.12.3 \\ \hline
\end{tabular}
\end{table}

The model training settings were meticulously tuned to ensure stable convergence across the beach litter dataset.
We define the input image resolution as $640 \times 640$ pixels to balance computational overhead and fine-grained detail retention.
The training process spanned 200 epochs with a batch size of 32. We utilized the Stochastic Gradient Descent (SGD) optimizer, where the initial learning rate was set to 0.01.
This configuration also incorporated a momentum factor of 0.937 and a weight decay of 0.0005 to prevent overfitting. Training hyper-parameters are summarized in Table ~\ref{tab:training_settings}.

\begin{table}[ht]
\centering
\caption{Configuration of model training hyper-parameters.}
\label{tab:training_settings}
\begin{tabular}{ll}
\hline
Hyper-parameter & Value \\ \hline
Input Image Resolution & $640 \times 640$ \\
Total Training Epochs & 200 \\
Batch Size & 32 \\
Optimizer & SGD \\
Initial Learning Rate & 0.01 \\
Momentum & 0.937 \\
Weight Decay & 0.0005 \\
Worker Threads & 25 \\ \hline
\end{tabular}
\end{table}

\subsection{Evaluation Criteria}

To quantitatively assess the detection and segmentation fidelity of the proposed model, we utilize standard performance metrics including precision, recall, and mean average precision (mAP), consistent with established protocols in environmental remote sensing \cite{scarricaNovelBeachLitter2022, zhaoApplicationImprovedMachine2024}.
The primary classification outcomes are categorized as follows:
\begin{itemize}
    \item \textbf{True Positive (TP)}: Debris instances correctly identified and segmented relative to the ground truth.
\item \textbf{False Positive (FP)}: Background features or incorrect categories erroneously identified as debris.
\item \textbf{False Negative (FN)}: Existing debris instances that the model fails to detect.
\end{itemize}

Precision and recall are calculated as follows:
\begin{equation}
\text{Precision} = \frac{\text{TP}}{\text{TP} + \text{FP}}
\end{equation}
\begin{equation}
\text{Recall} = \frac{\text{TP}}{\text{TP} + \text{FN}}
\end{equation}

The Intersection over Union (IoU) criterion is employed to determine the accuracy of the predicted masks.
For a given prediction, the IoU is defined as the ratio of the intersection area to the union area between the predicted mask ($M_p$) and the ground truth mask ($M_g$):
\begin{equation}
\text{IoU} = \frac{|M_p \cap M_g|}{|M_p \cup M_g|}
\end{equation}

Based on these calculated values, Average Precision (AP) is computed as the area under the precision-recall curve.
For multiple litter categories, we utilize the $mAP$, which is the mean of AP values across all categories at an IoU threshold of 0.5.
This metric serves as our primary indicator for evaluating model accuracy.
Furthermore, we assess computational efficiency using three key indicators: the number of learnable parameters (\text{Params}), and frames per second (\text{FPS}).
These metrics provide a comprehensive measure of the model's suitability for real-time deployment on UAV hardware.

\subsection{Benchmarking against Mainstream Segmentation Algorithms}

To evaluate the performance of the proposed framework, a comparative analysis was conducted against eleven instance segmentation models.
These benchmarks encompass three distinct architectural paradigms: (i) the transformer-based RTDETR-L-seg; (ii) the state-space-based Mamba-YOLO-T-seg;
and (iii) the lightweight convolutional lineage from YOLOv6s-seg to the baseline YOLOv26n-seg.
The quantitative performance metrics are summarized in Table ~\ref{tab:comparison}.

\begin{table}[htbp]
\centering
\caption{Performance comparison with mainstream instance segmentation algorithms on the beach litter dataset.
Precision ($P$), Recall ($R$), and mean Average Precision ($mAP$) are reported in percentages.}
\label{tab:comparison}
\resizebox{\linewidth}{!}{%
\begin{tabular}{lccccccc}
\toprule
Model & $P/\%$ & $R/\%$ & $mAP_{50}/\%$ & $mAP/\%$ & Params/M & Size/MB & FPS \\  
\midrule
RTDETR-L-seg & 61.2 & 45.9 & 50.8 & 41.8 & 30.81 & 63.8 & 143 \\
GhostNet-seg & 42.4 & 35.4 & 35.7 & 27.1 & 3.01 & 6.5 & 183 \\
Mamba-YOLO-T-seg & 27.2 & 58.7 & 49.4 & 39.6 & 6.23 & 12.8 & 63 \\
YOLOv6s-seg & 50.8 & 28.6 & 38.5 & 29.1 & 16.75 & 33.8 & 291 \\
YOLOv8n-seg & 45.4 & 38.0 & 38.6 & 30.3 & 3.26 
& 6.8 & 289 \\
YOLOv9s-seg & 40.9 & 49.9 & 50.5 & 41.3 & 8.52 & 18.0 & 107 \\
YOLOv10n-seg & 46.0 & 41.3 & 49.7 & 41.2 & 2.52 & 5.4 & 240 \\
YOLOv11n-seg & 52.1 & 44.3 & 52.6 & 43.9 & 2.84 & 6.0 & 232 \\
YOLOv12n-seg & 57.0 & 34.1 & 51.3 & 41.8 & 2.81 & 6.1 & 169 \\
YOLOv13n-seg & 33.9 & 50.1 & 50.4 & 42.0 & 2.72 & 6.0 & 117 \\
YOLOv26n-seg & 53.2 & 46.7 & 53.6 & 44.3 & 2.69 & 6.6 & 193 \\
\textbf{Ours} & \textbf{69.7} & \textbf{44.4} & 
\textbf{58.7} & \textbf{46.8} & \textbf{2.73} & \textbf{6.7} & \textbf{145} \\  
\bottomrule
\end{tabular}%
}
\end{table}

The comparative data reveal that PLAS-Net achieves a Precision of 69.7\%.
This high precision ensures that the data used for management decisions are grounded in actual anthropogenic inputs rather than background noise like shell fragments or vegetation.
Although this thresholding yields a slight decrease in overall recall to 44.4\%, such a trade-off is essential for coastal environments.
False positives derived from natural detritus would skew the downstream fragmentation models and area-based risk estimations.
The framework prioritizes the geometric fidelity of the extracted masks, establishing a reliable lower bound for subsequent quantification.
\begin{figure}[htbp]
    \centering
    \includegraphics[width=\linewidth]{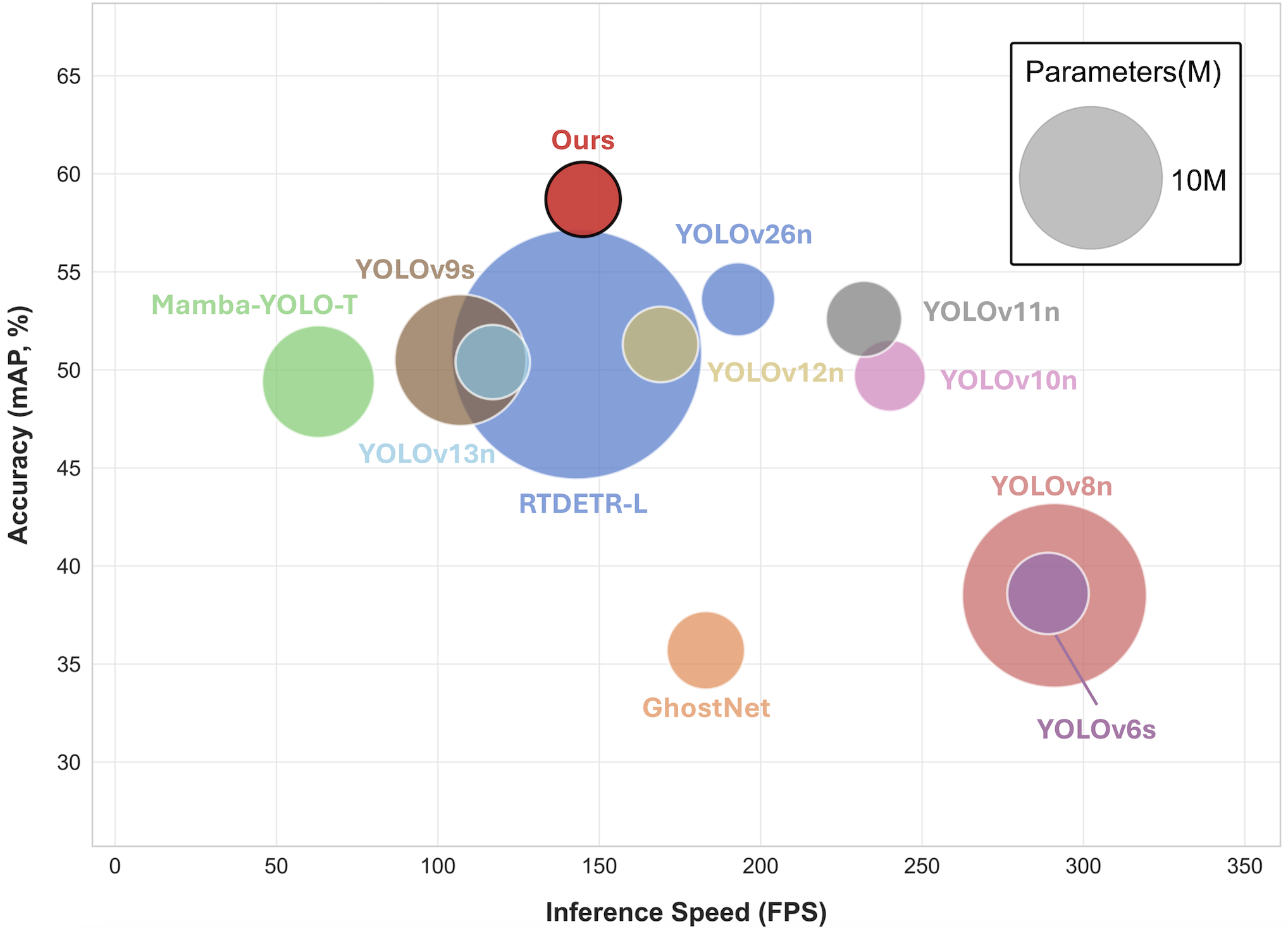}
    \caption{Comprehensive performance trade-off analysis of various instance segmentation models.
(Left) Model parameters (M) versus segmentation accuracy ($mAP$). (Right) Computational complexity (FLOPs) versus segmentation accuracy.
The radius of each bubble corresponds to the inference speed (FPS).
Our proposed framework achieves a competitive balance between segmentation accuracy, model size, and computational efficiency.}
    \label{fig:model_comparison}
\end{figure}

To intuitively illustrate the trade-off between segmentation accuracy and computational efficiency, the comprehensive performance metrics of the evaluated models are visualized in Fig. \ref{fig:model_comparison}.
As depicted, PLAS-Net is positioned in the upper-left quadrant of both the parameter and computational cost coordinate spaces.
While maintaining a highly competitive inference speed indicated by the bubble radius, our framework achieves the highest $mAP$ score with minimal parameters and FLOPs.
Compared to heavier architectures like RTDETR-L-seg and computationally intensive models like YOLOv6s-seg, the proposed network demonstrates a superior balance.
This combination of competitive accuracy and low parameter count suggests feasibility for deployment in near-real-time coastal monitoring workflows, pending validation on embedded hardware platforms.
\begin{figure}[htbp]
    \centering
    \includegraphics[width=\linewidth]{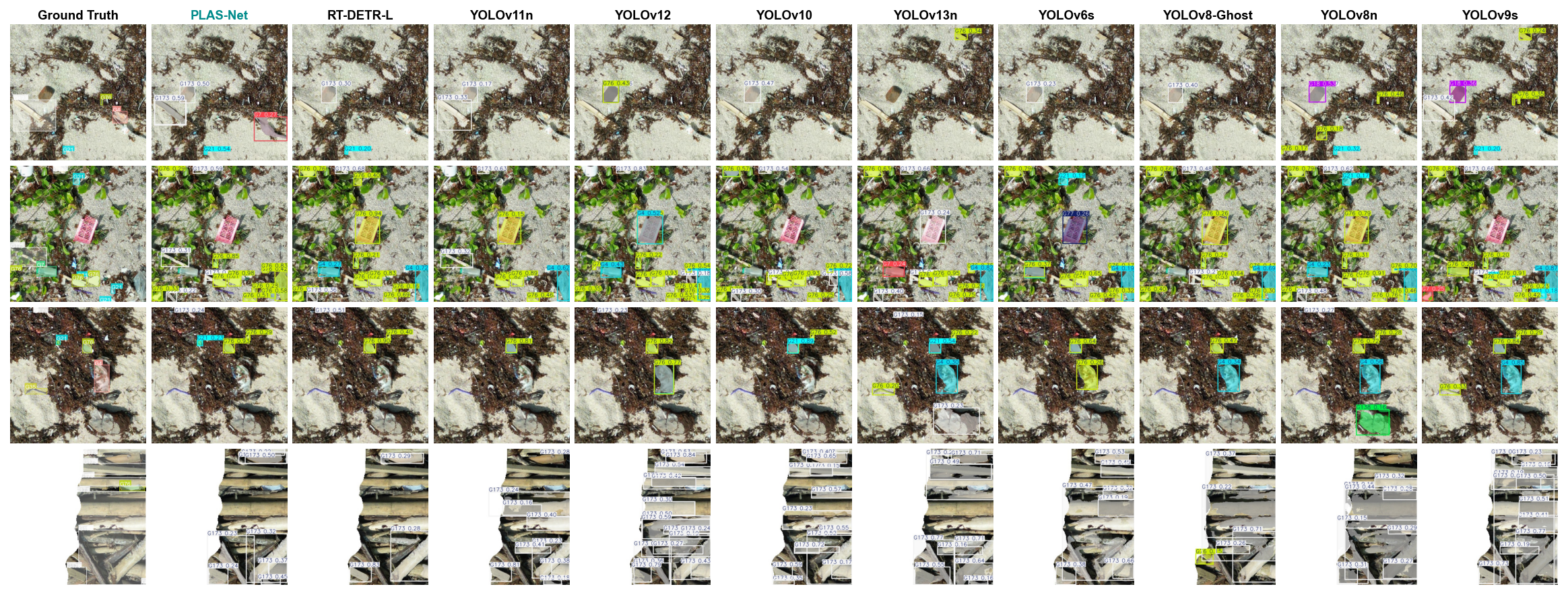}
    \caption{Visual comparison of detection and segmentation results across different models in complex beach environments.}
    \label{fig:comparison_visual}
\end{figure}

Visual evidence in Fig. \ref{fig:comparison_visual} further delineates these performance gaps.
In the first row, representing a low-contrast scenario with four ground truth targets, baseline models exhibit significant misclassification.
For instance, YOLOv8n and YOLOv9s-seg misidentify the central wood fragment (G-173) as a plastic basket (G-18).
While YOLOv13n and YOLOv11n detect only one and two valid targets respectively, PLAS-Net accurately locks all four objects.
In the second row, which presents a high-density scene with ten diverse targets, the complex vegetation background leads to extensive false negatives in mainstream models.
Specifically, while our model matches the ground truth count, YOLOv11n exhibits severe omissions due to the cluttered superposition of grass and sand.
In the remaining scenarios featuring dense wood remnants (G-173) and small plastic fragments (G-76), mainstream models often aggregate clustered litter into a single instance or fail to capture the features of small objects.
Our framework suppresses these omissions, maintaining consistent object number matching across all five representative cases.

\subsection{Ablation Analysis of Architectural Enhancements}

Ablation experiments were conducted to evaluate the contributions of C3kDFF, CPCAA, and DMSSF to pixel-level segmentation performance. The quantitative results are presented in Table~\ref{tab:ablation_results}.

\begin{table}[htbp]
\centering
\caption{Ablation results on the beach litter dataset.}
\label{tab:ablation_results}
\resizebox{\linewidth}{!}{%
\begin{tabular}{lcccccccccc}
\toprule
Model & C3kDFF & CPCAA & DMSSF & $P/\%$ & $R/\%$ & $mAP_{50}/\%$ & $mAP/\%$ & Params/M & Size/MB & FPS \\
\midrule
Baseline (YOLOv26n-seg)   & --         & --         & --         & 53.2 & 46.7 & 53.6 & 44.3 & 2.69 & 6.6 & 193 \\
Baseline + C3kDFF & \checkmark & --         & --         & 63.5 & 42.4 & 55.5 & 45.1 & 2.67 & 6.5 & 180 \\
Baseline + C3kDFF + CPCAA & \checkmark & \checkmark & --         & 65.4 & 46.8 & 57.8 & 45.9 & 2.66 & 6.5 & 179 \\
\textbf{PLAS-Net}  & \checkmark & \checkmark & \checkmark & \textbf{69.7} & \textbf{44.4} & \textbf{58.7} & \textbf{46.8} & \textbf{2.73} & \textbf{6.7} & \textbf{145} \\
\bottomrule
\end{tabular}%
}
\end{table}

\begin{figure}[htbp]
    \centering
    \includegraphics[width=\linewidth]{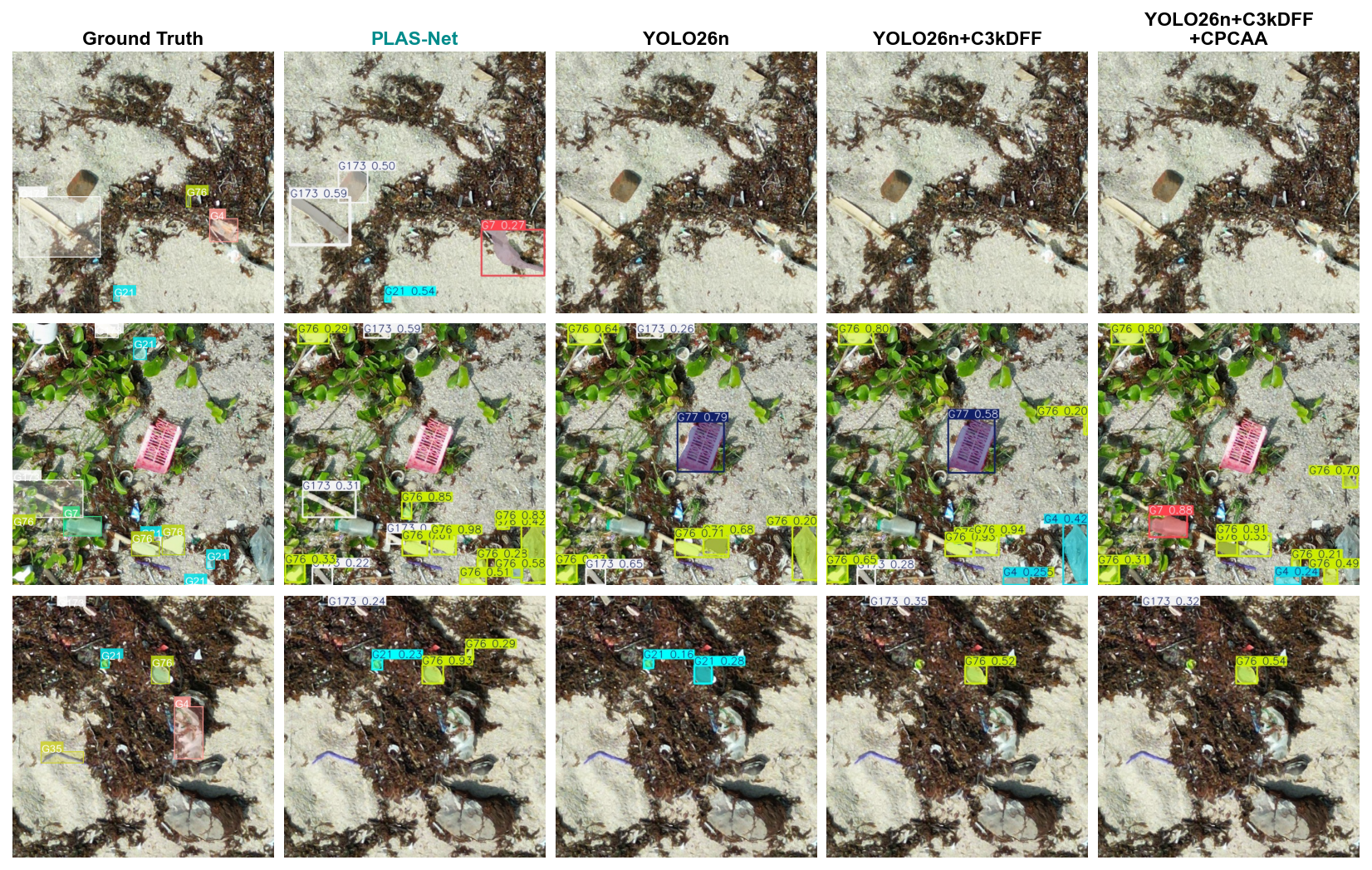}
    \caption{Visual comparison of ablation models across different scenarios.}
    \label{fig:ablation_visual}
\end{figure}

Fig.~\ref{fig:ablation_visual} shows how mask quality changes across models. In rows 1 and 2, where vegetation and gravel are present, the baseline model generates masks with boundary overflow. Grass and sand are included in predicted regions, showing that the model cannot separate debris from background textures. The full model produces compact masks that follow object boundaries closely. Even for partially buried objects such as G-173 wood fragments or G-21 plastic strips, the predicted regions remain constrained without enclosing surrounding materials. Rows 3 and 4 show fragmentation problems. The baseline and intermediate models generate scattered mask patches for small and weak objects such as G-76 plastic fragments, which do not form complete regions and are unstable under shadow or low contrast. The full model produces continuous masks that match object contours. In row 5, the task involves thin and elongated objects. The baseline models miss parts of these objects, producing incomplete and broken masks. The full model preserves the entire object extent with clear boundaries, even on high-texture sandy backgrounds.

\subsubsection{Effect of C3kDFF}

C3kDFF raises Precision from $53.2\%$ to $63.5\%$, reflecting a reduction in false positives. As shown in Fig.~\ref{fig:ablation_visual} (rows 1--2), the baseline model produces masks that extend into surrounding vegetation and sand. This is caused by static feature fusion, where shallow texture features and deep semantic features are combined without adaptive control. After C3kDFF is introduced, predicted regions become more compact and better aligned with true object areas, and background regions are effectively excluded. This improvement is achieved through dynamic feature fusion \parencite{yangDNetDynamicLarge2026}, which applies both channel-level and spatial-level calibration. Channel reweighting reduces the influence of irrelevant features, and spatial weighting highlights informative regions. Together, these operations help the model separate debris from visually similar background textures, reducing false detections.

\subsubsection{Effect of CPCAA}

CPCAA raises $mAP_{50}$ from $55.5\%$ to $57.8\%$, with improvement mainly in boundary quality and mask continuity. As shown in Fig.~\ref{fig:ablation_visual} (rows 3--4), without CPCAA the model produces fragmented masks where objects appear as multiple disconnected patches. This problem is most evident for small plastic fragments under shadow. With CPCAA, predicted masks become continuous and form a single connected region following the object shape. This change results from improved spatial context modeling \parencite{caiPolyKernelInception2024a}. Horizontal and vertical strip convolutions allow the network to capture long-range spatial information along object structures, maintaining continuity across weak or low-contrast regions. Mask fragmentation is reduced and boundary consistency improves.

\subsubsection{Effect of DMSSF}

DMSSF brings the model to its best performance, with $mAP_{50}$ of $58.7\%$ and $mAP$ of $46.8\%$, improving mask completeness and multi-scale consistency. As shown in Fig.~\ref{fig:ablation_visual} (rows 3--5), without DMSSF the model often produces incomplete masks, with parts of small or thin objects missing in complex sandy backgrounds. With DMSSF, predicted masks are complete and stable across object sizes. For thin and elongated objects in row 5, the full structure is preserved without missing parts. This improvement comes from modeling the relationship between feature maps at different scales \parencite{kangASFYOLONovelYOLO2024a}. Rather than treating each scale independently, the model learns a joint representation using 3D convolution. Dynamic upsampling further adjusts sampling positions based on feature content, retaining fine details during reconstruction \parencite{liuLearningUpsampleLearning2023}. These mechanisms ensure that both small objects and thin structures are fully represented in the pixel-level segmentation output.

\section{Discussion}

Having established that PLAS-Net produces more precise pixel-level masks than existing baselines, this section examines whether and how that improvement in mask quality translates into more reliable downstream environmental analyses.
Four model architectures are used as comparators throughout: YOLOv8n (a widely used convolutional baseline), Mamba-YOLO (state-space model), RT-DETR-L (transformer-based), and PLAS-Net.
Model evaluation metrics ($mAP_{50}$, precision, recall) reported in Section~3 were computed on the held-out test partition. For the downstream environmental analyses below, each trained model was deployed across the full orthomosaic to produce a scene-wide litter inventory, following standard remote sensing practice in which a validated model is applied as an inference tool over the entire study area \parencite{martinUseUnmannedAerial2018}. For each analysis, we ask whether mask fidelity affects the environmental conclusions drawn.
The subsequent subsections address three demonstration applications in turn: physical fragmentation dynamics modelled via NPD power-law fitting, spatial risk heterogeneity assessed through the ERI, and source composition patterns revealed by the abundance--area decoupling.
A final subsection outlines practical implications for coastal management workflows.

\subsection{Physical Degradation and Fragmentation Dynamics}

\begin{figure}[htbp]
    \centering
    \includegraphics[width=\textwidth, height=0.9\textheight, keepaspectratio]{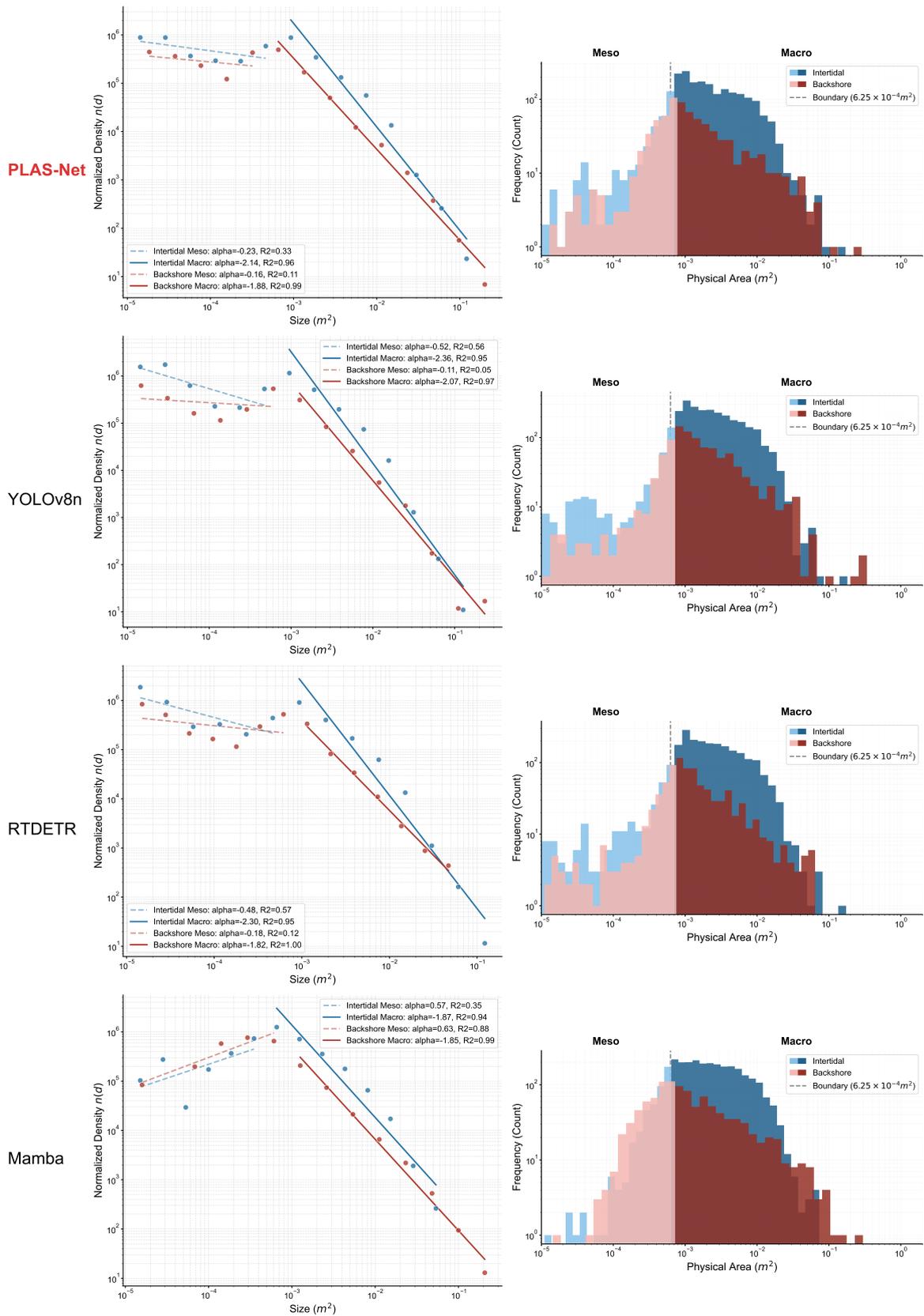}
    \caption{Comprehensive fragmentation analysis: (Left) NPD power-law fitting results and statistical significance;
(Right) Grain-size frequency analysis across different coastal zones. Among the evaluated models, PLAS-Net shows a closer alignment with the expected power-law pattern for scale-invariant fragmentation.}
    \label{fig:1a2_combined_fragmentation}
\end{figure}

Plastic pollution threatens global ecosystems through environmental persistence and continuous fragmentation, a process that exponentially increases bioavailability and ecological toxicity \parencite{macleodGlobalThreatPlastic2021}. Quantifying these fragmentation dynamics is essential for predicting the long-term fate of marine litter. However, existing remote sensing frameworks frequently overlook this physical dimension, predominantly focusing on algorithmic precision or abundance counts restricted by rectangular bounding boxes. This bounding box reliance precludes extracting the true planar areas required to model degradation. The transition from macro litter to secondary microplastics is a logarithmic cascade governed by the surface area-to-volume ratio and photochemical cleavage \parencite{Andrady2011, cozarPlasticDebrisOpen2014}. Because bounding boxes enclose substantial background regions and overestimate effective exposure surfaces, precise pixel-level decoupling via PLAS-Net is a prerequisite for authentic fragmentation analysis \parencite{martinUseUnmannedAerial2018}.

To bridge this gap, our study applies the NPD model, defining the scale-invariant fracture processes of marine plastics through power-law distributions \parencite{cozarPlasticDebrisOpen2014, SmithTurrell2021}. Grain size frequency analysis and fitting results demonstrate that PLAS-Net successfully restores physical continuity. The segmented power-law fit restricted to the Macro range ($A > 6.25 \times 10^{-4}$ m$^2$) yields scaling exponents of $\alpha = -2.14$ ($R^2 = 0.97$) in the Intertidal zone and $\alpha = -1.88$ ($R^2 = 0.99$) in the Backshore zone. These values fall between the theoretical expectations for two-dimensional surface fragmentation ($\alpha_{area} = -1.5$, corresponding to the breakup of sheet-like plastics) and the Cózar et al. (2014) steady-state volumetric fragmentation model ($\alpha_{area} = -2.0$). The intermediate position of the observed exponents is physically consistent with the mixed-material composition of the surveyed debris, which includes both quasi-planar items (plastic bags, sheeting) and three-dimensional objects (bottles, containers).

The validity of these fragmentation models is heavily dependent on observational resolution and material scope. When restricted to the Macro range, all four models achieve $R^2 > 0.93$ ($p < 0.001$), confirming that the power-law assumption holds above the GSD detection limit. In the Meso range below this threshold, all models exhibit $R^2 < 0.35$, consistent with the systematic sub-threshold truncation reported by \textcite{cozarPlasticDebrisOpen2014} for fragments below 5 mm in oceanic sampling. This detection-floor effect is instrument-inherent rather than model-specific, validating the restriction of power-law fitting to the Macro segment. Furthermore, the surveyed litter encompasses 14 G-code categories, of which four (wooden items, paper bags, other paper, and clothing) are non-polymeric materials. Although these collectively represent a minor fraction of detected instances (approximately 13.0\% of test-set objects), their inclusion means that the NPD fit characterises the fragmentation state of the mixed debris assemblage rather than of plastic polymers alone. Isolating polymer-only fragmentation dynamics would require supplementary material identification, for instance through hyperspectral sensing \parencite{garaba2018airborne}, which is beyond the scope of the current UAV-RGB workflow.

Notably, the Intertidal zone consistently yields steeper fragmentation exponents than the Backshore across all four evaluated models, contrary to the expectation that prolonged UV exposure in the Backshore would accelerate fragmentation. This reversal may reflect the dominance of mechanical abrasion by wave action and tidal cycling in the Intertidal zone, which can drive more rapid size reduction than photochemical degradation alone \parencite{Andrady2011}. These findings are consistent with the physical and geographical characteristics of the study area. As a semi-enclosed water body, the region experiences seasonal current reversals and intense wave-induced abrasion driven by monsoons \parencite{Yanagi2001_GoT_Monsoon, Phattananuruch2024_JMSE_GoT_FMD}. This consistency supports the view that extracting physical projection areas via instance segmentation can contribute to bridging computer vision methods and physical degradation modeling, providing a data foundation for evaluating environmental exposure \parencite{hidakaPixellevelImageClassification2022}.

\subsection{Refined Ecological Risk Assessment and Spatial Heterogeneity}

\begin{figure}[htbp]
    \centering
    \includegraphics[width=\textwidth, height=0.9\textheight, keepaspectratio]{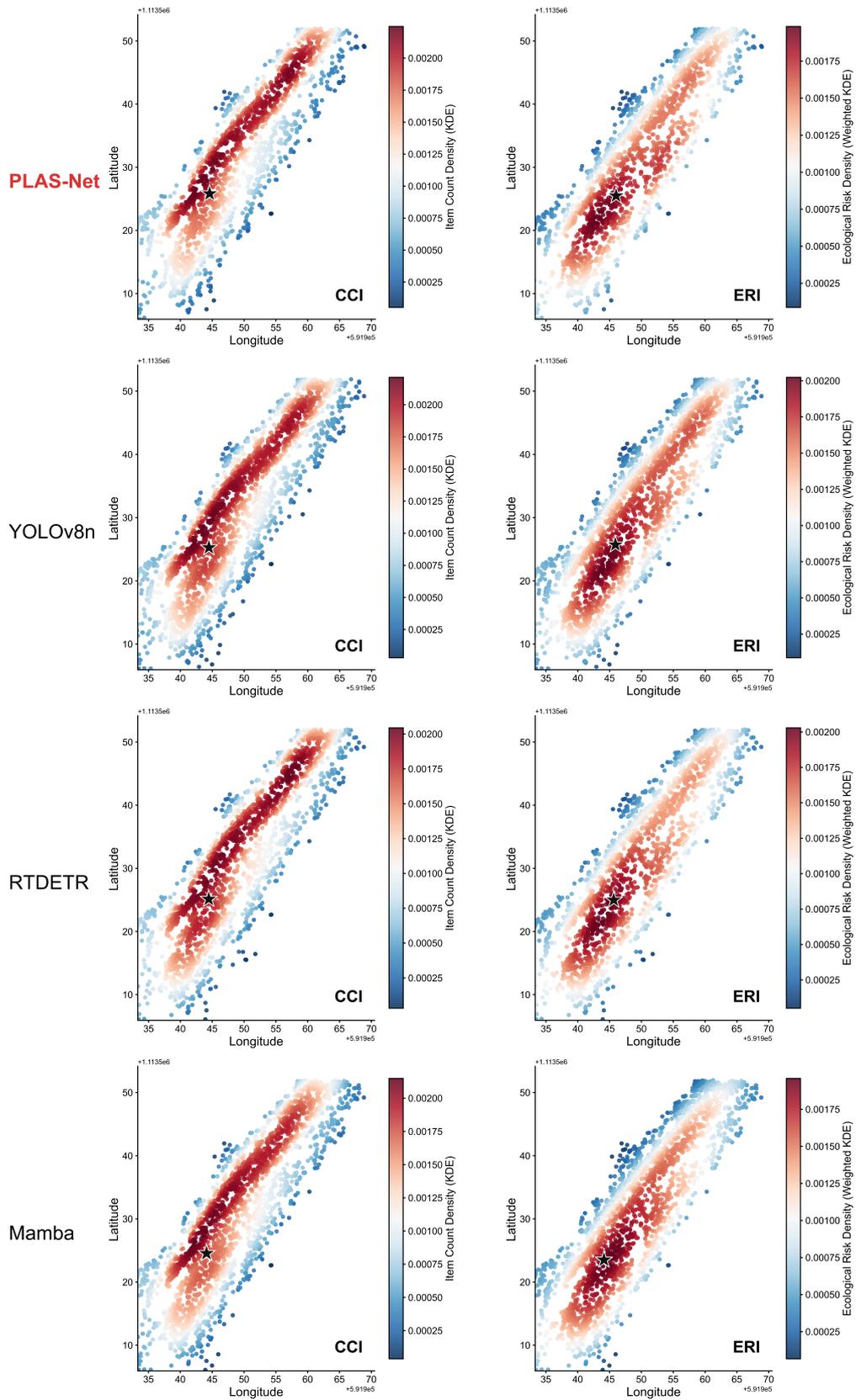}
    \caption{Spatial Pattern of Clean-Coast Index (CCI) and Ecological Risk Index (ERI).
PLAS-Net produces more spatially concentrated risk hotspot estimates, whereas baseline models (YOLOv8n and Mamba) tend to generate diffuse risk signals attributable to false-positive detections.
}
    \label{fig:3_detailed_step3}
\end{figure}

Finding where pollution gathers is the basis of coastal cleanup \parencite{jambeckPlasticWasteInputs2015a}. Uneven debris distribution means some local habitats face much higher exposure. Past work often focused only on improving algorithms. For example, \textcite{scarricaNovelBeachLitter2022} successfully generated pixel-level masks but did not translate them into environmental data. We must connect model outputs to actual conservation metrics to make these tools useful.

The ERI offers a clearer picture than the traditional CCI. The CCI simply counts objects. But how could a huge fishing net and a tiny cigarette butt contribute the same value to the index? Using physical area instead of counts removes this bias. We applied category-specific hazard weights ($w_k$, Table~\ref{tab:hazard_weights}) from \textcite{wilcoxUsingExpertElicitation2016} to account for different threat levels. A plastic bag ($w_k = 8$) and a paper bag ($w_k = 1$) might have the same size, but the plastic bag produces an ERI score eight times higher due to greater entanglement risks \parencite{wilcoxUsingExpertElicitation2016}. Consequently, locations with large and dangerous waste get higher scores. Counting items cannot reveal these high-risk zones.

We mapped both the CCI \parencite{alkalayCleancoastIndexNew2007} and the ERI \parencite{hakansonEcologicalRiskIndex1980} to compare the spatial risk patterns produced by different models (Fig.~\ref{fig:3_detailed_step3}). The choice of detection model changes the environmental conclusions entirely. PLAS-Net successfully locates concentrated risk hotspots. It identified a peak ERI of 5.8244 in sector S2. The baseline models fail to capture this pattern. YOLOv8n and Mamba spread high-risk signals across the entire beach. They overestimate the peak ERI by 25.2\% and 44.4\% respectively. These baselines produced over 1,000 false-positive detections. Adding these extra bounding boxes into the ERI calculation artificially inflates the risk values, and the resulting maps falsely suggest that extreme pollution covers the whole area uniformly. This error hides the actual layout of ecological threats. Ocean waves and currents naturally group debris together \parencite{preveniosBeachLitterDynamics2018}, so real pollution should appear patchy.

Finding the true accumulation zones is vital for managing Aow Luek Bay. This area in the Gulf of Thailand faces changing seasonal currents \parencite{Yanagi2001_GoT_Monsoon, Pokavanich2024_eGOT_Monsoon}. Regional winds and nearshore currents push large amounts of debris onto the shore \parencite{Phattananuruch2024_JMSE_GoT_FMD}. Also, the shape of the bay also traps this waste. Together, these physical forces cause trash to pile up heavily in specific spots instead of spreading evenly. Local cleanup records from groups like Roctopus ecoTrust confirm this uneven distribution. 

Counting items fails to capture the true danger to these habitats. Large corals like \textit{Porites lutea} and sponges suffer from tissue damage and algae overgrowth. Discarded fishing gear covering these organisms causes most of this harm \parencite{Pengsakun2026, Boerger2010, galganiMarineLitter2013}. One large fishing net smothers a much larger area of the reef than hundreds of small plastic fragments \parencite{Angiolillo2015, Consoli2018}. Relying on physical cover area offers a more accurate way to measure exposure and plan restoration efforts.

\subsection{Source-Sink Dynamics and the Abundance-Area Paradox}

\begin{figure}[htbp]
    \centering
    \includegraphics[width=\textwidth, height=0.9\textheight, keepaspectratio]{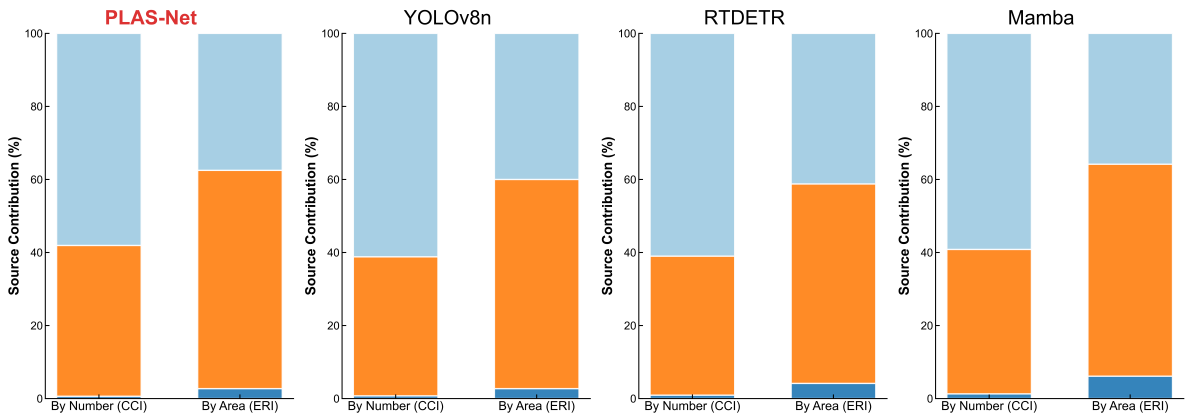}
    \caption{Comparison of source composition based on object abundance versus physical area across three functional origin categories. \textit{Domestic} comprises consumer-use items and packaging (G4, G6, G7, G21, G137, G138, G151, G158, G173); \textit{Fishing} comprises maritime and aquacultural gear (G18, G65); \textit{Fragments} comprises unidentifiable degraded plastic pieces (G76, G77). Left bars show percentage share by item count; right bars show percentage share by cumulative physical area ($\mathrm{m}^2$).}
    \label{fig:1c_source_sink}
\end{figure}

\begin{figure}[htbp]
    \centering
    \includegraphics[width=\textwidth, height=0.9\textheight, keepaspectratio]{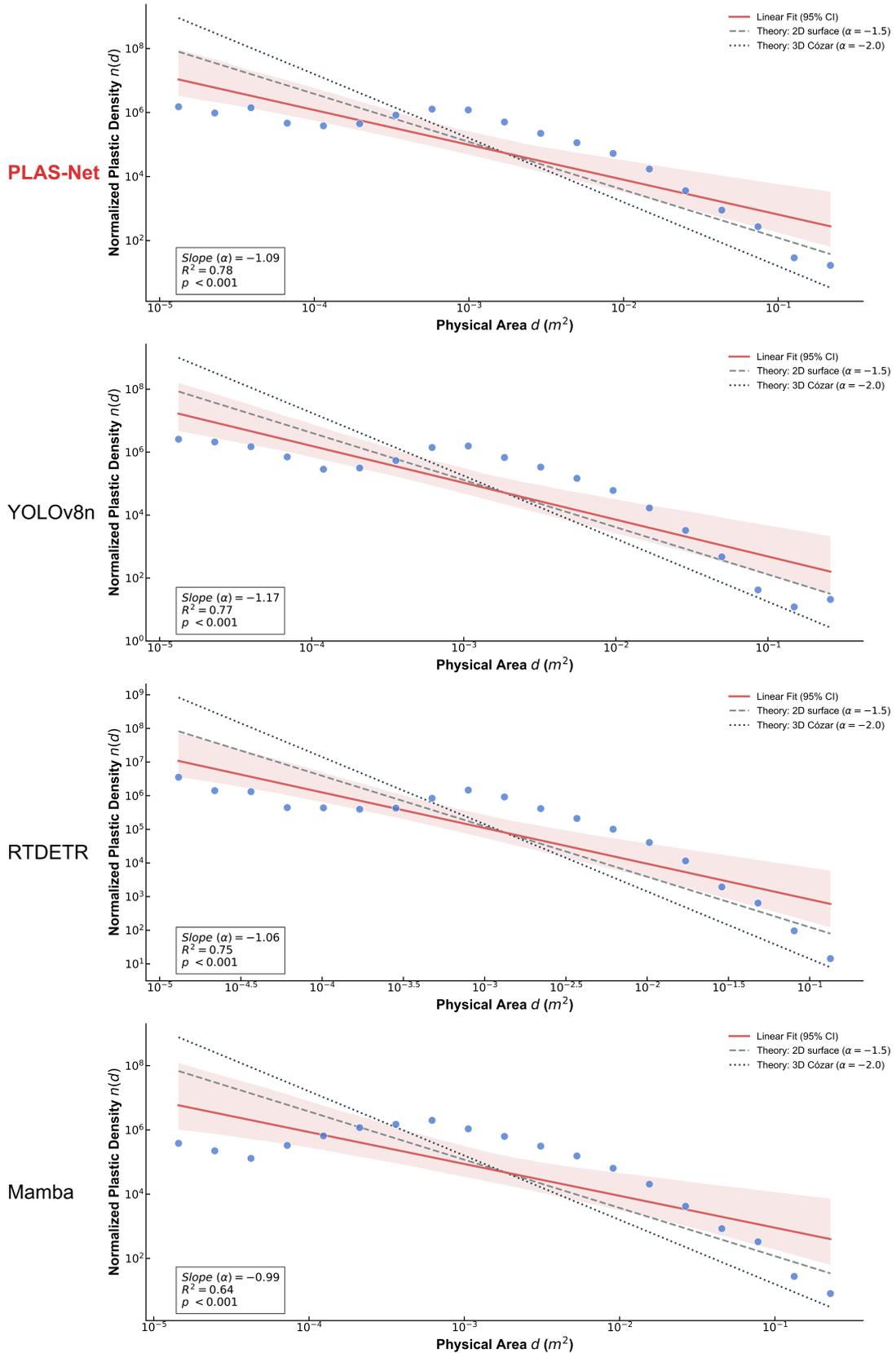}
    \caption{Class-level area distributions and the abundance-area inverse relationship across litter categories. Box plots show the per-item physical area ($\mathrm{m}^2$) distribution for each of the 14 G-code categories; the right panel illustrates the divergence between item count share and physical area share for the three source groups (Domestic, Fishing, Fragments).}
    \label{fig:4_detailed_source_sink}
\end{figure}

Marine plastic moves easily across borders \parencite{macleodGlobalThreatPlastic2021, jambeckPlasticWasteInputs2015}. Because of this mobility, coastal areas and remote islands usually collect waste from the ocean instead of generating it locally \parencite{barnesAccumulationFragmentationPlastic2009}. Finding the true origin of this debris means researchers must separate local pollution from distant accumulations \parencite{ryanMessageBottleAssessing2021}. This separation remains a necessary step for managing the global plastic crisis.

Traditional field surveys require too much effort and cover too little ground to map these source and sink relationships effectively. Uncrewed aerial vehicles offer an automated alternative. Still, most remote sensing studies focus on improving computer algorithms or just counting individual objects \parencite{kakoEstimationPlasticMarine2020, martinUseUnmannedAerial2018}. Methods that rely on bounding box detection can distort our understanding of where pollution comes from. Over time, one large plastic object breaks into hundreds of smaller fragments \parencite{cozarPlasticDebrisOpen2014}. Simply counting items makes the fragment category look larger than it actually is. This counting bias hides the original source of the waste.

To overcome this problem, physical area data must be extracted at the pixel level. Instance segmentation achieves this precision by separating true object boundaries from the background. Better segmentation accuracy helps build reliable models of pollution sources and sinks. It is a practical necessity for ecology rather than just a technical exercise.

Our framework captures an inverse relationship between item count and physical area (Fig. \ref{fig:4_detailed_source_sink}). Domestic items appear less often but have a larger average size of 0.00589 m\textsuperscript{2} per item in our model. At the same time, unidentifiable fragments dominate the total count but have a much smaller average area of 0.00263 m\textsuperscript{2}. This distribution aligns with the progressive breakdown of larger objects in place \parencite{cozarPlasticDebrisOpen2014}. Geometry alone cannot reveal the exact travel history of each item. Even so, the overall pattern clearly reflects this ongoing degradation.

Baseline models utilizing bounding boxes introduce severe quantitative biases during this physical translation.
YOLOv8n and Mamba overestimate fragment abundance by 41.6 percent and 36.9 percent, respectively.
These additional detections are false positives generated by background noise and coastal vegetation.
Such algorithmic over-detection fundamentally corrupts the source-sink modeling by artificially exaggerating local fragmentation rates.
The disparity between object count and physical footprint becomes most pronounced in fishing-related debris. In PLAS-Net, fishing gear (G18 and G65 combined) accounts for only 0.64\% of total item count yet contributes 2.73\% of cumulative physical area, representing a fourfold overrepresentation relative to abundance (Fig.~\ref{fig:4_detailed_source_sink}). Although the absolute area share is modest, the per-item average area of fishing gear (0.0127~m\textsuperscript{2}) is 4.8$\times$ that of unidentifiable fragments (0.00263~m\textsuperscript{2}), confirming that individual fishing items impose disproportionate spatial footprints despite their low frequency.
By isolating geometric morphology, we establish a signature of a transboundary pollution sink via PLAS-Net.
The accumulation of these massive spatial footprints is deeply influenced by regional hydrodynamics.
During the northeast monsoon, surface transport in the Gulf of Thailand is dominated by onshore winds and nearshore currents that push pelagic, industrial scale fishing gear toward the coast \parencite{Guo2021_GoT_NEM_Circulation, Phattananuruch2024_JMSE_GoT_FMD}.
Furthermore, coastal features like Aow Luek Bay function as topographic traps, amplifying the retention of these high mass floating materials \parencite{CritchellLambrechts2016_ECSS_PlasticAccumulation, Pokavanich2024}.
The presence of massive fishing gear, which requires significant ocean current energy to transport, confirms a fluid dynamics driven decoupling of local discard from regional accumulation.
Evaluating this transboundary flux through area based geometric quantification prevents statistical misattribution, offering evidence for regional marine conservation strategies.

\subsection{Practical Applications: A Proposed Multi-Tiered Coastal Management Workflow}
The combination of affordable consumer-grade UAVs and an improved instance segmentation model provides a practical basis for a multi-tiered coastal management system. At a strategic level, recognizing discarded fishing gear as a primary transboundary footprint allows policy to move from reactive beach cleanups toward proactive offshore interception \parencite{ryanMessageBottleAssessing2021}. These spatial results support the use of strict gear marking regulations to prevent maritime pollution, which aligns with existing warnings about benthic degradation in the Gulf of Thailand \parencite{pengsakunEnvironmentalEffectsPlastic2026}.

On a accuracy level, the model generates accurate risk hotspot maps by filtering out algorithmic false positives. These maps highlight topographic traps created by monsoon currents where plastic naturally accumulates \parencite{Phattananuruch2024_JMSE_GoT_FMD, CritchellLambrechts2016_ECSS_PlasticAccumulation}. NGOs operating with limited budgets can use this data to optimize where they send volunteers. Deploying teams with such precision ensures that resources are not wasted in clean areas and that debris removal is maximized in heavy accumulation centers.

Detailed monitoring is also possible by using instance segmentation to extract physical areas and track the NPD scaling exponent \parencite{macleodGlobalThreatPlastic2021}. This process functions as an early warning system for ecological threats. When the exponent approaches the steady-state fragmentation threshold of $\alpha_{area} = -2.0$, it indicates that macro-plastics are fracturing due to long-term weathering. These figures can trigger local alerts, allowing for preemptive removal before the plastic breaks down further.

\subsection{Methodological Limitations and Future Perspectives}
This study establishes a framework for monitoring coasts, but several limitations exist. Testing was conducted at a single beach to prove the initial concept. While the model handled complex sand and vegetation well, researchers must validate the system on other coastal types like mudflats or mangroves to ensure it works broadly. Because the training and testing data come from the same image map, shared traits like lighting and sand texture might lead to higher accuracy scores. These metrics may change if the model is used at an entirely independent site.

The current analysis relies on data from a single point in time. Future research should include surveys across different seasons to track how debris accumulates and breaks down over longer periods. This framework also uses two-dimensional footprints to estimate physical decay. Although pixel-level measurements are more accurate than simple bounding boxes, they cannot provide three-dimensional volume. 

A 2D area may underestimate the environmental impact of buried or irregular objects. Future work should incorporate UAV-based LiDAR or stereo photogrammetry to reconstruct 3D volumes. Using these technologies will allow for more precise mass estimation and better risk assessments.

\section{Conclusion}

PLAS-Net represents an instance segmentation framework that integrates the C3kDFF, CPCAA, and DMSSF modules to address visual interferences common in coastal environments. On the beach litter dataset, the framework produced a $mAP_{50}$ of 58.7\% with an inference speed of 145 FPS. This result reflects a balance between segmentation accuracy and computational efficiency, establishing a foundation for automated coastal monitoring.
 
The shift from object detection to instance segmentation provides pixel-level masks rather than bounding boxes. These masks enable downstream applications in physical modeling and environmental assessment. In practice, geometric quantification derived from masks improved the characterization of spatial heterogeneity in pollution distribution across a tropical pocket beach case study. High-fidelity masks proved more accurate than traditional bounding boxes, which tend to overestimate debris area. The masks allow application of NPD and ERI models to measure real-world environmental impacts.
 
Instance segmentation outputs translate into quantitative environmental indicators for coastal monitoring and management. The framework separates object abundance from physical footprints, offering a scalable tool to guide targeted cleanup interventions. This capacity to provide accurate spatial information supports evidence-based marine pollution policies.

\section*{Declaration of Competing Interest}
The authors confirm that there are no known financial or personal conflicts of interest that could have influenced the objective presentation of the findings in this research.

\section*{Acknowledgements}
The authors would like to thank the New Heaven Reef Conservation Program for their support during the field study in Thailand. We are grateful to Mr. Mowen Zhang, and Miss Yulun Chen from Southwest Forestry University for their dedicated assistance with the scientific activities and their constructive feedback during the manuscript preparation. 

This research was supported by the Japan Science and Technology Agency (JST) through the SPRING Program (Grant Number: JPMJSP2108) and the PRESTO Program (Grant Number: JPMJPR24G9).
% =========================================================
% Bibliography
% =========================================================
\bibliographystyle{unsrt} 
\bibliography{references}

@Article{rs16193617,
AUTHOR = {Sozio, Angelo and Scarrica, Vincenzo Mariano and Rizzo, Angela and Aucelli, Pietro Patrizio Ciro and Barracane, Giovanni and Dimuccio, Luca Antonio and Ferreira, Rui and La Salandra, Marco and Staiano, Antonino and Tarantino, Maria Pia and Scicchitano, Giovanni},
TITLE = {Application of Direct and Indirect Methodologies for Beach Litter Detection in Coastal Environments},
JOURNAL = {Remote Sensing},
VOLUME = {16},
YEAR = {2024},
NUMBER = {19},
ARTICLE-NUMBER = {3617},
URL = {https://www.mdpi.com/2072-4292/16/19/3617},
ISSN = {2072-4292},
DOI = {10.3390/rs16193617}
}

@Article{su14148311,
AUTHOR = {Song, Kyounghwan and Jung, Jung-Yeul and Lee, Seung Hyun and Park, Sanghyun and Yang, Yunjung},
TITLE = {Assessment of Marine Debris on Hard-to-Reach Places Using Unmanned Aerial Vehicles and Segmentation Models Based on a Deep Learning Approach},
JOURNAL = {Sustainability},
VOLUME = {14},
YEAR = {2022},
NUMBER = {14},
ARTICLE-NUMBER = {8311},
URL = {https://www.mdpi.com/2071-1050/14/14/8311},
ISSN = {2071-1050},
DOI = {10.3390/su14148311}
}

@article{FALLATI2019133581,
title = {Anthropogenic Marine Debris assessment with Unmanned Aerial Vehicle imagery and deep learning: A case study along the beaches of the Republic of Maldives},
journal = {Science of The Total Environment},
volume = {693},
pages = {133581},
year = {2019},
issn = {0048-9697},
doi = {https://doi.org/10.1016/j.scitotenv.2019.133581},
url = {https://www.sciencedirect.com/science/article/pii/S0048969719335065},
author = {L. Fallati and A. Polidori and C. Salvatore and L. Saponari and A. Savini and P. Galli}
}

@article{alkalayCleancoastIndexNew2007,
  title = {Clean-Coast Index---{{A}} New Approach for Beach Cleanliness Assessment},
  author = {Alkalay, Ronen and Pasternak, Galia and Zask, Alon},
  year = 2007,
  month = jan,
  journal = {Ocean \& Coastal Management},
  volume = {50},
  number = {5},
  pages = {352--362},
  issn = {0964-5691},
  doi = {10.1016/j.ocecoaman.2006.10.002},
  urldate = {2026-03-27}
}

@article{barnesAccumulationFragmentationPlastic2009,
  title = {Accumulation and Fragmentation of Plastic Debris in Global Environments},
  author = {Barnes, David K. A. and Galgani, Francois and Thompson, Richard C. and Barlaz, Morton},
  year = 2009,
  month = jul,
  journal = {Philosophical Transactions of the Royal Society B: Biological Sciences},
  volume = {364},
  number = {1526},
  pages = {1985--1998},
  issn = {0962-8436},
  doi = {10.1098/rstb.2008.0205},
  urldate = {2026-01-07}
}

@article{barryTop10Marine2023,
  title = {Top 10 Marine Litter Items on the Seafloor in {{European}} Seas from 2012 to 2020},
  author = {Barry, Jon and Rindorf, Anna and Gago, Jesus and Silburn, Briony and McGoran, Alex and Russell, Josie},
  year = 2023,
  month = dec,
  journal = {Science of The Total Environment},
  volume = {902},
  pages = {165997},
  issn = {0048-9697},
  doi = {10.1016/j.scitotenv.2023.165997},
  urldate = {2026-01-21}
}

@inproceedings{caiPolyKernelInception2024a,
  title = {Poly {{Kernel Inception Network}} for {{Remote Sensing Detection}}},
  booktitle = {2024 {{IEEE}}/{{CVF Conference}} on {{Computer Vision}} and {{Pattern Recognition}} ({{CVPR}})},
  author = {Cai, Xinhao and Lai, Qiuxia and Wang, Yuwei and Wang, Wenguan and Sun, Zeren and Yao, Yazhou},
  year = 2024,
  month = jun,
  pages = {27706--27716},
  issn = {2575-7075},
  doi = {10.1109/CVPR52733.2024.02617},
  urldate = {2026-03-26}
}

@article{chengMethodsDatasetsSemantic2024,
  title = {Methods and Datasets on Semantic Segmentation for {{Unmanned Aerial Vehicle}} Remote Sensing Images: {{A}} Review},
  shorttitle = {Methods and Datasets on Semantic Segmentation for {{Unmanned Aerial Vehicle}} Remote Sensing Images},
  author = {Cheng, Jian and Deng, Changjian and Su, Yanzhou and An, Zeyu and Wang, Qi},
  year = 2024,
  month = may,
  journal = {ISPRS Journal of Photogrammetry and Remote Sensing},
  volume = {211},
  pages = {1--34},
  issn = {0924-2716},
  doi = {10.1016/j.isprsjprs.2024.03.012},
  urldate = {2025-07-06}
}

@misc{ClassifyPlasticWaste,
  title = {Classify Plastic Waste as Hazardous \textbar{} {{Nature}}},
  urldate = {2026-01-07},
  howpublished = {https://www.nature.com/articles/494169a}
}

@misc{galgani2023guidance,
  title={Guidance on the Monitoring of Marine Litter in European Seas: An update to improve the harmonised monitoring of marine litter under the Marine Strategy Framework Directive},
  author={Galgani, Francois and Ruiz-Orej{\'o}n, LF and Ronchi, Francesca and Tallec, Kevin and Fischer, Elke and Matiddi, Marco and Anastasopoulou, Aikaterini and Andresmaa, Eva and Angiolillo, Michela and Bakker, Michael Paiva and others},
  year={2023},
  publisher={Publications office of the European Union}
}

@article{garaba2018airborne,
  title={An airborne remote sensing case study of synthetic hydrocarbon detection using short wave infrared absorption features identified from marine-harvested macro-and microplastics},
  author={Garaba, Shungudzemwoyo P and Dierssen, Heidi M},
  journal={Remote Sensing of Environment},
  volume={205},
  pages={224--235},
  year={2018},
  publisher={Elsevier}
}

@article{cozarPlasticDebrisOpen2014,
  title = {Plastic Debris in the Open Ocean},
  author = {C{\'o}zar, Andr{\'e}s and Echevarr{\'i}a, Fidel and {Gonz{\'a}lez-Gordillo}, J. Ignacio and Irigoien, Xabier and {\'U}beda, B{\'a}rbara and {Hern{\'a}ndez-Le{\'o}n}, Santiago and Palma, {\'A}lvaro T. and Navarro, Sandra and {Garc{\'i}a-de-Lomas}, Juan and Ruiz, Andrea and {Fern{\'a}ndez-de-Puelles}, Mar{\'i}a L. and Duarte, Carlos M.},
  year = 2014,
  month = jul,
  journal = {Proceedings of the National Academy of Sciences},
  volume = {111},
  number = {28},
  pages = {10239--10244},
  issn = {0027-8424, 1091-6490},
  doi = {10.1073/pnas.1314705111},
  urldate = {2026-03-27},
  langid = {english}
}

@article{CritchellLambrechts2016_ECSS_PlasticAccumulation,
  title = {Modelling Accumulation of Marine Plastics in the Coastal Zone; What Are the Dominant Physical Processes?},
  author = {Critchell, Kay and Lambrechts, Jonathan},
  year = 2016,
  month = mar,
  journal = {Estuarine, Coastal and Shelf Science},
  volume = {171},
  pages = {111--122},
  issn = {0272-7714},
  doi = {10.1016/j.ecss.2016.01.036},
  urldate = {2026-01-19}
}

@article{Guo2021_GoT_NEM_Circulation,
  title = {Thermohaline Conditions and Circulation in the {{Gulf}} of {{Thailand}} during the Northeast Monsoon},
  author = {Guo, Jingsong and Qu, Dapeng and Zhang, Zhixin and Sangmanee, Chalermrat and Chanthasiri, Nuttida and Guo, Binghuo},
  year = 2021,
  month = aug,
  journal = {Continental Shelf Research},
  volume = {225},
  pages = {104487},
  issn = {0278-4343},
  doi = {10.1016/j.csr.2021.104487},
  urldate = {2026-01-18}
}

@article{hakansonEcologicalRiskIndex1980,
  title = {An Ecological Risk Index for Aquatic Pollution Control.a Sedimentological Approach},
  author = {Hakanson, Lars},
  year = 1980,
  month = jan,
  journal = {Water Research},
  volume = {14},
  number = {8},
  pages = {975--1001},
  issn = {0043-1354},
  doi = {10.1016/0043-1354(80)90143-8},
  urldate = {2026-03-27}
}

@article{hidakaPixellevelImageClassification2022,
  title = {Pixel-Level Image Classification for Detecting Beach Litter Using a Deep Learning Approach},
  author = {Hidaka, Mitsuko and Matsuoka, Daisuke and Sugiyama, Daisuke and Murakami, Koshiro and Kako, Shin'ichiro},
  year = 2022,
  month = feb,
  journal = {Marine Pollution Bulletin},
  volume = {175},
  pages = {113371},
  issn = {0025-326X},
  doi = {10.1016/j.marpolbul.2022.113371},
  urldate = {2026-01-07}
}

@article{jambeckPlasticWasteInputs2015,
  title = {Plastic Waste Inputs from Land into the Ocean},
  author = {Jambeck, Jenna R. and Geyer, Roland and Wilcox, Chris and Siegler, Theodore R. and Perryman, Miriam and Andrady, Anthony and Narayan, Ramani and Law, Kara Lavender},
  year = 2015,
  month = feb,
  journal = {Science},
  volume = {347},
  number = {6223},
  pages = {768--771},
  publisher = {American Association for the Advancement of Science},
  doi = {10.1126/science.1260352},
  urldate = {2026-03-26}
}

@article{jambeckPlasticWasteInputs2015a,
  title = {Plastic Waste Inputs from Land into the Ocean},
  author = {Jambeck, Jenna R. and Geyer, Roland and Wilcox, Chris and Siegler, Theodore R. and Perryman, Miriam and Andrady, Anthony and Narayan, Ramani and Law, Kara Lavender},
  year = 2015,
  month = feb,
  journal = {Science},
  volume = {347},
  number = {6223},
  pages = {768--771},
  publisher = {American Association for the Advancement of Science},
  doi = {10.1126/science.1260352},
  urldate = {2026-03-27}
}

@article{kakoEstimationPlasticMarine2020,
  title = {Estimation of Plastic Marine Debris Volumes on Beaches Using Unmanned Aerial Vehicles and Image Processing Based on Deep Learning},
  author = {Kako, Shin'ichiro and Morita, Shohei and Taneda, Tetsuya},
  year = 2020,
  month = jun,
  journal = {Marine Pollution Bulletin},
  volume = {155},
  pages = {111127},
  issn = {0025-326X},
  doi = {10.1016/j.marpolbul.2020.111127},
  urldate = {2026-03-27}
}

@article{kangASFYOLONovelYOLO2024a,
  title = {{{ASF-YOLO}}: {{A}} Novel {{YOLO}} Model with Attentional Scale Sequence Fusion for Cell Instance Segmentation},
  shorttitle = {{{ASF-YOLO}}},
  author = {Kang, Ming and Ting, Chee-Ming and Ting, Fung Fung and Phan, Rapha{\"e}l C. -W.},
  year = 2024,
  month = jul,
  journal = {Image and Vision Computing},
  volume = {147},
  pages = {105057},
  issn = {0262-8856},
  doi = {10.1016/j.imavis.2024.105057},
  urldate = {2026-03-26}
}

@inproceedings{liuLearningUpsampleLearning2023,
  title = {Learning to {{Upsample}} by {{Learning}} to {{Sample}}},
  booktitle = {Proceedings of the {{IEEE}}/{{CVF International Conference}} on {{Computer Vision}}},
  author = {Liu, Wenze and Lu, Hao and Fu, Hongtao and Cao, Zhiguo},
  year = 2023,
  pages = {6027--6037},
  urldate = {2026-03-26},
  langid = {english}
}

@article{lopez-arquilloInterdependenceCoastalTourist2024,
  title = {Interdependence in {{Coastal Tourist Territories}} between {{Marine Litter}} and {{Immediate Tourist Zoning Density}}: {{Methodological Approach}} for {{Urban Sustainable Development}}},
  shorttitle = {Interdependence in {{Coastal Tourist Territories}} between {{Marine Litter}} and {{Immediate Tourist Zoning Density}}},
  author = {{L{\'o}pez-Arquillo}, Juan Diego and Oliveira, Cristiana and Serrano Gonz{\'a}lez, Jose and Dur{\'a}n S{\'a}nchez, Amador},
  year = 2024,
  month = jan,
  journal = {Land},
  volume = {13},
  number = {1},
  pages = {50},
  publisher = {Multidisciplinary Digital Publishing Institute},
  issn = {2073-445X},
  doi = {10.3390/land13010050},
  urldate = {2025-12-08},
  langid = {english}
}

@article{macleodGlobalThreatPlastic2021,
  title = {The Global Threat from Plastic Pollution},
  author = {MacLeod, Matthew and Arp, Hans Peter H. and Tekman, Mine B. and Jahnke, Annika},
  year = 2021,
  month = jul,
  journal = {Science},
  volume = {373},
  number = {6550},
  pages = {61--65},
  issn = {0036-8075, 1095-9203},
  doi = {10.1126/science.abg5433},
  urldate = {2026-03-27},
  langid = {english}
}

@article{martinUseUnmannedAerial2018,
  title = {Use of Unmanned Aerial Vehicles for Efficient Beach Litter Monitoring},
  author = {Martin, Cecilia and Parkes, Stephen and Zhang, Qiannan and Zhang, Xiangliang and McCabe, Matthew F. and Duarte, Carlos M.},
  year = 2018,
  month = jun,
  journal = {Marine Pollution Bulletin},
  volume = {131},
  pages = {662--673},
  issn = {0025-326X},
  doi = {10.1016/j.marpolbul.2018.04.045},
  urldate = {2026-01-07}
}

@article{pengsakunEnvironmentalEffectsPlastic2026,
  title = {Environmental Effects of Plastic Pollution from Lost, Discarded, and Abandoned Fishing Gear on Underwater Pinnacles in the {{Gulf}} of {{Thailand}}},
  author = {Pengsakun, Sittiporn and Yeemin, Thamasak and Sutthacheep, Makamas and Jungrak, Laongdow and Klinthong, Wanlaya and Aunkhongthong, Wiphawan and Chamchoy, Charernmee and Sukkeaw, Maneerat and Odthon, Saowalak and Suebpala, Wichin},
  year = 2026,
  month = jan,
  journal = {Frontiers in Marine Science},
  volume = {12},
  publisher = {Frontiers},
  issn = {2296-7745},
  doi = {10.3389/fmars.2025.1670284},
  urldate = {2026-03-26},
  langid = {english}
}

@article{Phattananuruch2024_JMSE_GoT_FMD,
  title = {Monsoon-{{Driven Dispersal}} of {{River-Sourced Floating Marine Debris}} in {{Tropical Semi-Enclosed Waters}}: {{A Case Study}} in the {{Gulf}} of {{Thailand}}},
  shorttitle = {Monsoon-{{Driven Dispersal}} of {{River-Sourced Floating Marine Debris}} in {{Tropical Semi-Enclosed Waters}}},
  author = {Phattananuruch, Kittipong and Pokavanich, Tanuspong and Phattananuruch, Kittipong and Pokavanich, Tanuspong},
  year = 2024,
  month = dec,
  journal = {Journal of Marine Science and Engineering},
  volume = {12},
  number = {12},
  publisher = {publisher},
  issn = {2077-1312},
  doi = {10.3390/jmse12122258},
  urldate = {2026-01-19},
  langid = {english}
}

@article{Pokavanich2024_eGOT_Monsoon,
  title = {Seasonal {{Dynamics}} and {{Three-Dimensional Hydrographic Features}} of the {{Eastern Gulf}} of {{Thailand}}: {{Insights}} from {{High-Resolution Modeling}} and {{Field Measurements}}},
  shorttitle = {Seasonal {{Dynamics}} and {{Three-Dimensional Hydrographic Features}} of the {{Eastern Gulf}} of {{Thailand}}},
  author = {Pokavanich, Tanuspong and Worrawatanathum, Vasawan and Phattananuruch, Kittipong and Koolkalya, Sontaya and Pokavanich, Tanuspong and Worrawatanathum, Vasawan and Phattananuruch, Kittipong and Koolkalya, Sontaya},
  year = 2024,
  month = jul,
  journal = {Water},
  volume = {16},
  number = {14},
  publisher = {publisher},
  issn = {2073-4441},
  doi = {10.3390/w16141962},
  urldate = {2026-01-18},
  langid = {english}
}

@article{preveniosBeachLitterDynamics2018,
  title = {Beach Litter Dynamics on {{Mediterranean}} Coasts: {{Distinguishing}} Sources and Pathways},
  shorttitle = {Beach Litter Dynamics on {{Mediterranean}} Coasts},
  author = {Prevenios, Michael and Zeri, Christina and Tsangaris, Catherine and Liubartseva, Svitlana and Fakiris, Elias and Papatheodorou, George},
  year = 2018,
  month = apr,
  journal = {Marine Pollution Bulletin},
  volume = {129},
  number = {2},
  pages = {448--457},
  issn = {0025-326X},
  doi = {10.1016/j.marpolbul.2017.10.013},
  urldate = {2026-01-20}
}

@article{ryanMessageBottleAssessing2021,
  title = {Message in a Bottle: {{Assessing}} the Sources and Origins of Beach Litter to Tackle Marine Pollution},
  shorttitle = {Message in a Bottle},
  author = {Ryan, Peter G. and Weideman, Eleanor A. and Perold, Vonica and Hofmeyr, Greg and Connan, Ma{\"e}lle},
  year = 2021,
  month = nov,
  journal = {Environmental Pollution},
  volume = {288},
  pages = {117729},
  issn = {0269-7491},
  doi = {10.1016/j.envpol.2021.117729},
  urldate = {2026-01-07}
}

@article{ryanMonitoringAbundancePlastic2009,
  title = {Monitoring the Abundance of Plastic Debris in the Marine Environment},
  author = {Ryan, Peter G. and Moore, Charles J. and {van Franeker}, Jan A. and Moloney, Coleen L.},
  year = 2009,
  month = jul,
  journal = {Philosophical Transactions of the Royal Society B: Biological Sciences},
  volume = {364},
  number = {1526},
  pages = {1999--2012},
  issn = {0962-8436},
  doi = {10.1098/rstb.2008.0207},
  urldate = {2026-01-07}
}

@misc{sapkotaYOLO26KeyArchitectural2026,
  title = {{{YOLO26}}: {{Key Architectural Enhancements}} and {{Performance Benchmarking}} for {{Real-Time Object Detection}}},
  shorttitle = {{{YOLO26}}},
  author = {Sapkota, Ranjan and Cheppally, Rahul Harsha and Sharda, Ajay and Karkee, Manoj},
  year = 2026,
  month = mar,
  number = {arXiv:2509.25164},
  eprint = {2509.25164},
  primaryclass = {cs},
  publisher = {arXiv},
  doi = {10.48550/arXiv.2509.25164},
  urldate = {2026-03-20},
  archiveprefix = {arXiv}
}

@article{scarricaNovelBeachLitter2022,
  title = {A Novel Beach Litter Analysis System Based on {{UAV}} Images and {{Convolutional Neural Networks}}},
  author = {Scarrica, Vincenzo M. and Aucelli, Pietro P. C. and Cagnazzo, Cosimo and Casolaro, Angelo and Fiore, Pierpaolo and La Salandra, Marco and Rizzo, Angela and Scardino, Giovanni and Scicchitano, Giovanni and Staiano, Antonino},
  year = 2022,
  month = dec,
  journal = {Ecological Informatics},
  volume = {72},
  pages = {101875},
  issn = {1574-9541},
  doi = {10.1016/j.ecoinf.2022.101875},
  urldate = {2026-01-07}
}

@article{smithMonitoringPlasticBeach2021,
  title = {Monitoring {{Plastic Beach Litter}} by {{Number}} or by {{Weight}}: {{The Implications}} of {{Fragmentation}}},
  shorttitle = {Monitoring {{Plastic Beach Litter}} by {{Number}} or by {{Weight}}},
  author = {Smith, Lauren and Turrell, William Richard},
  year = 2021,
  month = sep,
  journal = {Frontiers in Marine Science},
  volume = {8},
  publisher = {Frontiers},
  issn = {2296-7745},
  doi = {10.3389/fmars.2021.702570},
  urldate = {2025-12-08},
  langid = {english}
}

@article{songComparativeStudyDeep2021,
  title = {A Comparative Study of Deep Learning-Based Network Model and Conventional Method to Assess Beach Debris Standing-Stock},
  author = {Song, Kyounghwan and Jung, Jung-Yeul and Lee, Seung Hyun and Park, Sanghyun},
  year = 2021,
  month = jul,
  journal = {Marine Pollution Bulletin},
  volume = {168},
  pages = {112466},
  issn = {0025-326X},
  doi = {10.1016/j.marpolbul.2021.112466},
  urldate = {2025-12-10}
}

@article{steinmetz-weissMappingDroneApplications2025,
  title = {Mapping {{Drone Applications}} in {{Rural}} and {{Regional Cities}}: {{A Scoping Review}} of the {{Australian State}} of {{Practice}}},
  shorttitle = {Mapping {{Drone Applications}} in {{Rural}} and {{Regional Cities}}},
  author = {{Steinmetz-Weiss}, Christine and Marshall, Nancy and Bishop, Kate and Wei, Yuan},
  year = 2025,
  month = jan,
  journal = {Applied Sciences},
  volume = {15},
  number = {15},
  pages = {8519},
  publisher = {Multidisciplinary Digital Publishing Institute},
  issn = {2076-3417},
  doi = {10.3390/app15158519},
  urldate = {2025-12-10},
  langid = {english}
}

@article{vansebillePhysicalOceanographyTransport2020,
  title = {The Physical Oceanography of the Transport of Floating Marine Debris},
  author = {{van Sebille}, Erik and Aliani, Stefano and Law, Kara Lavender and Maximenko, Nikolai and Alsina, Jos{\'e} M and Bagaev, Andrei and Bergmann, Melanie and Chapron, Bertrand and Chubarenko, Irina and C{\'o}zar, Andr{\'e}s and Delandmeter, Philippe and Egger, Matthias and {Fox-Kemper}, Baylor and Garaba, Shungudzemwoyo P and {Goddijn-Murphy}, Lonneke and Hardesty, Britta Denise and Hoffman, Matthew J and Isobe, Atsuhiko and Jongedijk, Cleo E and Kaandorp, Mikael L A and Khatmullina, Liliya and Koelmans, Albert A and Kukulka, Tobias and Laufk{\"o}tter, Charlotte and Lebreton, Laurent and Lobelle, Delphine and Maes, Christophe and {Martinez-Vicente}, Victor and Morales Maqueda, Miguel Angel and {Poulain-Zarcos}, Marie and Rodr{\'i}guez, Ernesto and Ryan, Peter G and Shanks, Alan L and Shim, Won Joon and Suaria, Giuseppe and Thiel, Martin and {van den Bremer}, Ton S and Wichmann, David},
  year = 2020,
  month = feb,
  journal = {Environmental Research Letters},
  volume = {15},
  number = {2},
  pages = {023003},
  publisher = {IOP Publishing},
  issn = {1748-9326},
  doi = {10.1088/1748-9326/ab6d7d},
  urldate = {2025-12-08},
  langid = {english}
}

@article{wilcoxUsingExpertElicitation2016,
  title = {Using Expert Elicitation to Estimate the Impacts of Plastic Pollution on Marine Wildlife},
  author = {Wilcox, Chris and Mallos, Nicholas J. and Leonard, George H. and Rodriguez, Alba and Hardesty, Britta Denise},
  year = 2016,
  month = mar,
  journal = {Marine Policy},
  volume = {65},
  pages = {107--114},
  issn = {0308-597X},
  doi = {10.1016/j.marpol.2015.10.014},
  urldate = {2025-12-10}
}

@article{Yanagi2001_GoT_Monsoon,
  title = {Seasonal {{Variation}} of {{Stratification}} in the {{Gulf}} of {{Thailand}}},
  author = {Yanagi, Tetsuo and Sachoemar, Suhendar I. and Takao, Toshiyuki and Fujiwara, Shunji},
  year = 2001,
  month = aug,
  journal = {Journal of Oceanography},
  volume = {57},
  number = {4},
  pages = {461--470},
  issn = {1573-868X},
  doi = {10.1023/A:1021237721368},
  urldate = {2026-01-18},
  langid = {english}
}

@article{yangDNetDynamicLarge2026,
  title = {D-{{Net}}: {{Dynamic}} Large Kernel with Dynamic Feature Fusion for Volumetric Medical Image Segmentation},
  shorttitle = {D-{{Net}}},
  author = {Yang, Jin and Qiu, Peijie and Zhang, Yichi and Marcus, Daniel S. and Sotiras, Aristeidis},
  year = 2026,
  month = mar,
  journal = {Biomedical Signal Processing and Control},
  volume = {113},
  pages = {108837},
  issn = {1746-8094},
  doi = {10.1016/j.bspc.2025.108837},
  urldate = {2026-03-26}
}

@article{zhaoApplicationImprovedMachine2024,
  title = {Application of Improved Machine Learning in Large-Scale Investigation of Plastic Waste Distribution in Tourism {{Intensive}} Artificial Coastlines},
  author = {Zhao, Haoluan and Wang, Xiaoli and Yu, Xun and Peng, Shitao and Hu, Jianbo and Deng, Mengtao and Ren, Lijun and Zhang, Xiaodan and Duan, Zhenghua},
  year = 2024,
  month = sep,
  journal = {Environmental Pollution},
  volume = {356},
  pages = {124292},
  issn = {0269-7491},
  doi = {10.1016/j.envpol.2024.124292},
  urldate = {2026-01-07}
}

@article{zouHighQualityInstanceSegmentationNetwork2022,
  title = {A {{High-Quality Instance-Segmentation Network}} for {{Floating-Algae Detection Using RGB Images}}},
  author = {Zou, Yibo and Wang, Xiaoliang and Wang, Lei and Chen, Ke and Ge, Yan and Zhao, Linlin},
  year = 2022,
  month = dec,
  journal = {Remote Sensing},
  volume = {14},
  number = {24},
  publisher = {publisher},
  issn = {2072-4292},
  doi = {10.3390/rs14246247},
  urldate = {2026-01-21},
  langid = {english}
}

@article{Angiolillo2015,
  title={Distribution and assessment of marine debris in the deep Tyrrhenian Sea (NW Mediterranean Sea, Italy)},
  author={Angiolillo, Michela and di Lorenzo, Barbara and Farcomeni, Alessio and Bo, Marzia and Bavestrello, Giorgio and Santangelo, Giovanni and Cau, Alessandro and Cobianchi, Maria and de Rossi, Flora and Canese, Simone},
  journal={Marine Pollution Bulletin},
  volume={92},
  number={1-2},
  pages={149--159},
  year={2015},
  publisher={Elsevier}
}

@article{Consoli2018,
  title={Marine litter in submarine canyons of the Mediterranean Sea},
  author={Consoli, Pierpaolo and Romeo, Teresa and Angiolillo, Michela and Canese, Simone and Esposito, Valentina and Salvati, Eva and Scotti, Gianfranco and Andaloro, Franco},
  journal={Marine Pollution Bulletin},
  volume={128},
  pages={42--51},
  year={2018},
  publisher={Elsevier}
}

@article{Pengsakun2026,
  title={Environmental effects of plastic pollution from lost, discarded, and abandoned fishing gear on underwater pinnacles in the Gulf of Thailand},
  author={Pengsakun, S. and Yeemin, T. and Sutthacheep, M. and Jungrak, L. and Klinthong, W. and Aunkhongthong, W. and Chamchoy, C. and Sukkeaw, M. and Odthon, S. and Suebpala, W.},
  journal={Frontiers in Marine Science},
  volume={12},
  pages={1670284},
  year={2026},
  doi={10.3389/fmars.2025.1670284}
}

@article{Andrady2011,
  title={Microplastics in the marine environment},
  author={Andrady, Anthony L.},
  journal={Marine Pollution Bulletin},
  volume={62},
  number={8},
  pages={1596--1605},
  year={2011},
  publisher={Elsevier}
}

@article{SmithTurrell2021,
  title={Monitoring plastic beach litter by number or by weight: the implications of fragmentation},
  author={Smith, S. D. and Turrell, W. R.},
  journal={Frontiers in Marine Science},
  volume={8},
  pages={702570},
  year={2021},
  publisher={Frontiers Media SA}
}

@article{Boerger2010,
  title={Plastic ingestion by planktivorous fishes in the North Pacific Central Gyre},
  author={Boerger, Christiana M. and Lattin, Gwendolyn L. and Moore, Charles J. and Kassapple, Chelsea J.},
  journal={Marine Pollution Bulletin},
  volume={60},
  number={12},
  pages={2275--2278},
  year={2010},
  publisher={Elsevier}
}

@article{galganiMarineLitter2013,
  title={Marine litter within the European Marine Strategy Framework Directive},
  author={Galgani, Fran{\c{c}}ois and Hanke, Georg and Werner, Stefanie and Oosterbaan, Lex and Nilsson, Per and Fleet, David and Kinsey, Susan and Thompson, Richard C. and Van Franeker, Jan and Vlachogianni, Thomais},
  journal={ICES Journal of Marine Science},
  volume={70},
  number={6},
  pages={1055--1064},
  year={2013},
  publisher={Oxford University Press}
}

@article{Pokavanich2024,
  title={Coastal hydrodynamic and plastic debris trajectory modeling in the semi-enclosed Gulf of Thailand},
  author={Pokavanich, Tanuspong and others},
  journal={Marine Pollution Bulletin},
  volume={198},
  pages={115800}, 
  year={2024},
  publisher={Elsevier}
}

@article{liu2026water,
  title = {Water-aware real-time detection of floating plastic debris via an enhanced YOLOv13 framework for aquatic pollution monitoring},
  author = {Liu, Zunyu and Wang, Jiao and Wu, Hao and Xue, Feng and Qin, Zixiang and Sun, Shan and Guo, Xianglong and others},
  journal = {Expert Systems with Applications},
  volume = {313},
  pages = {131552},
  year = {2026},
  doi = {10.1016/j.eswa.2026.131552},
  publisher={Elsevier}
}

@article{zhao2024riverbed,
  title={Riverbed litter monitoring using consumer-grade aerial-aquatic speedy scanner (AASS) and deep learning based super-resolution reconstruction and detection network},
  author={Zhao, Fan and Liu, Yongying and Wang, Jiaqi and Chen, Yijia and Xi, Dianhan and Shao, Xinlei and others},
  journal={Marine Pollution Bulletin},
  volume={209},
  pages={117030},
  year={2024},
  doi={10.1016/j.marpolbul.2024.117030},
  publisher={Elsevier}
}

@article{zhao2025seafloor,
  title={Seafloor debris detection using underwater images and deep learning-driven image restoration: A case study from Koh Tao, Thailand},
  author={Zhao, Fan and Huang, Baoxi and Wang, Jiaqi and Shao, Xinlei and Wu, Qingyang and Xi, Dianhan and others},
  journal={Marine Pollution Bulletin},
  volume={214},
  pages={117710},
  year={2025},
  doi={10.1016/j.marpolbul.2025.117710},
  publisher={Elsevier}
}

@article{tao2025diffusion,
  title={Diffusion-Enhanced Underwater Debris Detection via Improved YOLOv12n Framework},
  author={Tao, Jianghan and Zhao, Fan and Chen, Yijia and Liu, Yongying and Xue, Feng and Song, Jian and others},
  journal={Remote Sensing},
  volume={17},
  number={23},
  pages={3910},
  year={2025},
  publisher={MDPI}
}

@article{zhao2025novel,
  title={A novel underwater Holothurians monitoring system using consumer-grade amphibious UAV with Mamba-based Super-Resolution Reconstruction and enhanced YOLOv10},
  author={Zhao, Fan and Shao, Xinlei and Wang, Jiaqi and Chen, Yijia and Xi, Dianhan and Liu, Yongying and others},
  journal={Marine Environmental Research},
  pages={107510},
  year={2025},
  doi={10.1016/j.marenvres.2025.107510},
  publisher={Elsevier}
}

@article{zhao2026cost,
  title={Cost-effective ecological monitoring in shallow waters using amphibious unmanned aerial vehicles (AUAV) and deep learning-based computer vision},
  author={Zhao, Fan and Wang, Jiaqi and Chen, Yijia and Shao, Xinlei and Liu, Yongying and Chen, Yulun and others},
  journal={Marine Environmental Research},
  volume={216},
  pages={107911},
  year={2026},
  doi={10.1016/j.marenvres.2026.107911},
  publisher={Elsevier}
}

@article{Maximenko2019_IMDOS,
  title={Toward the integrated marine debris observing system},
  author={Maximenko, Nikolai and Corradi, Paolo and Law, Kara Lavender and Van Sebille, Erik and Garaba, Shungudzemwoyo P and Lampitt, Richard S and Galgani, Francois and Martinez-Vicente, Victor and Goddijn-Murphy, Lonneke and Veiga, Joana M and others},
  journal={Frontiers in Marine Science},
  volume={6},
  pages={447},
  year={2019},
  publisher={Frontiers Media SA},
  doi={10.3389/fmars.2019.00447}
}

\end{document}